%% file: arxiv_acl_latex.tex
\pdfoutput=1

\documentclass[11pt]{article}

\usepackage[final]{acl}

\usepackage{times}
\usepackage{latexsym}

\usepackage[T1]{fontenc}

\usepackage[utf8]{inputenc}

\usepackage{microtype}

\usepackage{inconsolata}

\usepackage{graphicx}

\usepackage{hyperref}
\usepackage{float}
\usepackage{url}
\usepackage{amsmath}
\usepackage{tikz}
\usepackage{sidecap}
\usepackage{wrapfig}
\usepackage{xcolor}

\let\svthefootnote\thefootnote
\newcommand\freefootnote[1]{%
  \let\thefootnote\relax%
  \footnotetext{#1}%
  \let\thefootnote\svthefootnote%
}
\title{Language models and brains align due to more than next-word prediction and word-level information}

\author{Gabriele Merlin \\
MPI for Software Systems \\
  Saarbrücken, Germany \\
  \texttt{gmerlin@mpi-sws.org} \\\And
  Mariya Toneva \\
  MPI for Software Systems \\ Saarbrücken, Germany \\
  \texttt{mtoneva@mpi-sws.org} \\}

\begin{document}
\maketitle
\freefootnote{Code available at 

\href{https://github.com/gab709/brain-llm-beyond-next-word}{\texttt{github.com/gab709/brain-llm-beyond-next-word}}.}

\begin{abstract}
Pretrained language models have been shown to significantly predict brain recordings of people comprehending language. Recent work suggests that the prediction of the next word is a key mechanism that contributes to this alignment. What is not yet understood is whether prediction of the next word is necessary for this observed alignment or simply sufficient, and whether there are other shared mechanisms or information that are similarly important. In this work, we take a step towards understanding the reasons for brain alignment via two simple perturbations in popular pretrained language models. These perturbations help us design contrasts that can control for different types of information. By contrasting the brain alignment of these differently perturbed models, we show that improvements in alignment with brain recordings are due to more than improvements in next-word prediction and word-level information.
\end{abstract}

\section{Introduction}\label{sec:introduction}
Language models (LMs) that have been pretrained to predict the next word over billions of text documents have also been shown to significantly predict brain recordings of people comprehending language \citep{wehbe2014aligning,jain2018incorporating,toneva2019interpreting,caucheteux2020language,schrimpf2021neural,goldstein2022shared}. Understanding the reasons behind the observed similarities between representations of language in machines and representations of language in the brain can lead to more insight into both systems. Recent studies suggest that the prediction of the next word is a key mechanism that contributes to the alignment between the two \citep{goldstein2022shared}. What is not yet understood is whether prediction of the next word is necessary for this observed alignment or simply sufficient, and whether other shared information is similarly important.

\begin{figure}[t]
\centering
\includegraphics[width=0.47\textwidth]{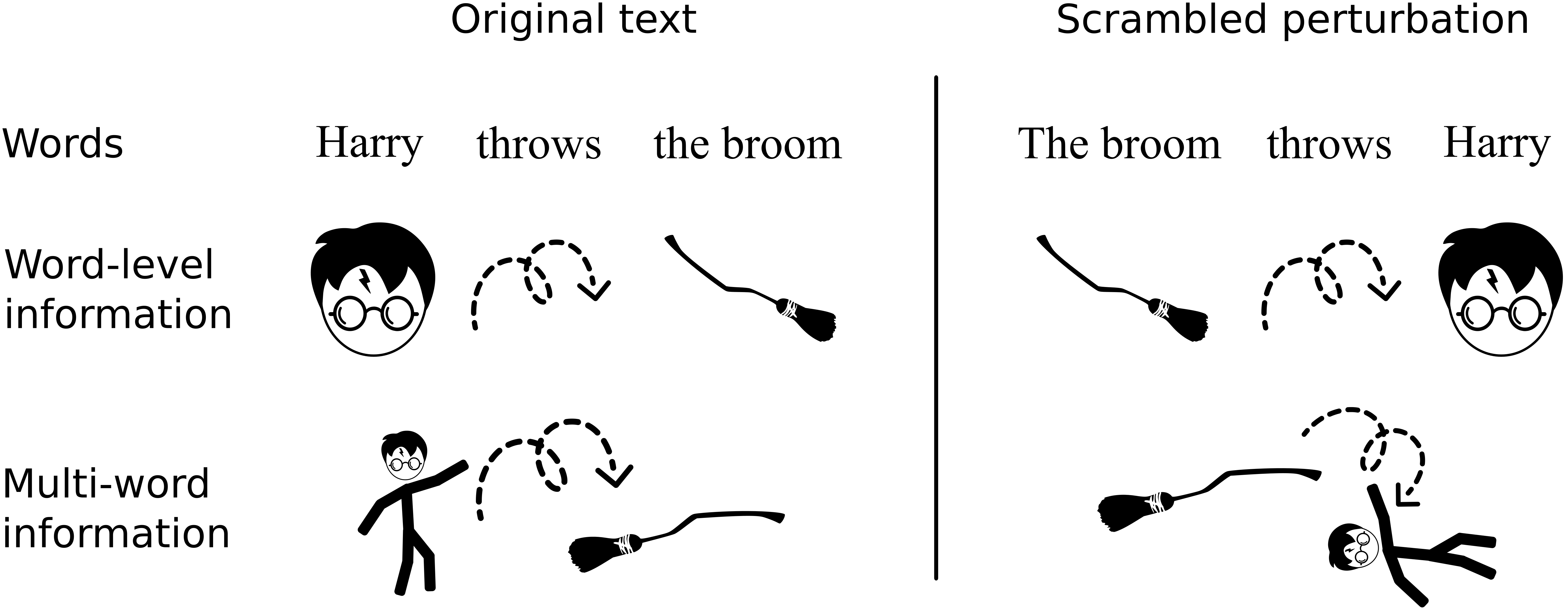}
\caption{An illustration of additional information that may be important for alignment between language models and brain recordings. Our approach is largely agnostic about the exact linguistic information contained in the conceptual quantities "word-level information" and "multi-word information", and the only assumption is that "word-level information" is not affected by word order.}
  \label{fig:linguistic_information}
\end{figure}

Understanding the impact of other kinds of information on brain alignment is complicated by correlations with next-word prediction (NWP). Because NWP is the LM training objective, better NWP
may also be related to improved representations of other types of information \citep{piantadosi2022meaning} that the human brain is sensitive to, such as word-level or multi-word information \citep{lerner2011topographic}. Neuroscientists are still investigating the exact linguistic features at the word- and multi-word levels that are important for processing in the brain, so for the remainder of the paper, we take an abstract approach and refer to “word-level information” as the non-contextualized representation of the word, and to “multi-word information” as relating to multiple words (e.g. syntax). For example, each word in “Harry throws the broom” has a non-contextualized meaning and the phrase has a different meaning depending on the word order (“Harry throws the broom” vs. “The broom throws Harry”, see Figure \ref{fig:linguistic_information}).
We note that these are conceptual quantities and not ones we are claiming to be able to quantify directly. The only assumption key to our argument is that word-level information is not impacted by word order. Other than that, our methods are agnostic about the specific linguistic information contained by these conceptual quantities. Both word-level and multi-word information may contribute to brain alignment, but their effect cannot be disentangled from that of next-word prediction using previous approaches.

In this work, we aim to disentangle the contributions of next-word prediction and word-level information from other factors, such as multi-word information, in the brain alignment of GPT-2-based models \citep{radford2019language}. 

Our methodology builds upon the traditional neuroscientific approach of constructing \emph{contrasts} between brain activity elicited by different conditions. A contrast reveals the processing of a specific property \emph{P} in the brain, by subtracting the brain activity elicited by two conditions (condition \emph{A} and \emph{B}) that are tightly controlled to contain similar information except for the target property \emph{P}. We leverage this approach and design contrasts between the \emph{predicted} brain activity by two related models: an original model and its perturbed version.  
By contrasting the brain alignment of these two models, we can conclude that any difference in brain alignment is due to the perturbation. Conversely, if the two contrasted conditions are controlled for some factor, then any difference in the \emph{predicted} brain activity between the two conditions cannot be due to this factor.

Our key insight is to design a contrast that controls for both information related to next-word prediction and word-level information. This contrast is enabled by two proposed perturbations. The first perturbation, which we name \emph{input scrambling}, scrambles the order of the input words at inference time. This perturbation controls for the word-level information when contrasting a model's brain alignment related to the original vs. the scrambled inputs, because, by definition, the word-level information encoded in the model representations remains the same. Any remaining brain alignment after the contrast must therefore be due to factors beyond word-level information, such as next-word prediction or multi-word information. The second perturbation further disentangles the contribution of next-word prediction to brain alignment. This perturbation, which we name \emph{stimulus-tuning}, fine-tunes a model to predict the next word in the specific naturalistic stimulus text corresponding to the brain recordings. The fine-tuning is done until the next-word prediction performance matches a pre-defined level above its input-scrambled version, similar to the baseline model's improvement over its input-scrambled version. Contrasting brain predictions from pairs of models—-baseline vs. scrambled and stimulus-tuned vs. scrambled—-controls for both word-level information and next-word prediction. Any residual brain alignment is then due to other factors, such as multi-word information.
We note that our methodology is based on models trained to predict the next word; however, we only control for next-word prediction performance at inference time. Therefore, we cannot draw conclusions about the importance of the training objective itself, but rather about the information that has emerged after the training process.

After controlling for word-level and next-word prediction in the final contrast, we still observe residual brain alignment. Across three types of models (GPT-2-small, GPT-2-medium \citep{radford2019language} and GPT-2-distilled \citep{sanh2019distilbert}), we find consistent residual brain alignment in two specific brain areas that are thought to process language \citep{fedorenko2010new,fedorenko2014reworking}--the inferior frontal gyrus (IFG) and the angular gyrus (AG)--suggesting that the brain alignment between the language model and these brain regions is due to more than next-word prediction and word-level information. We speculate that this alignment is due to multi-word information, which is consistent with previous findings about processing in these regions \citep{friederici2012cortical,humphreys2021unifying}.

Our main contributions are as follows:
(i) propose perturbations to pretrained language models that, when combined in suitable contrasts, can control for the effects of next-word prediction and word-level information on brain alignment; (ii) demonstrate that a proposed perturbation, which consist of tuning a language model on a validation stimulus text, can increase the alignment with brain recordings that correspond to a heldout text; (iii) reveal that the brain alignment with language regions, in particular in the inferior frontal gyrus (IFG) and the angular gyrus (AG), is due to more than next-word prediction and word-level information.

\section{Methods}\label{sec:methods}

\subsection{Baseline models}\label{subsec:methods_baseline}
We use GPT-2-based language models \citep{radford2019language} as the baseline pretrained language models. In particular, we investigate GPT-2-small, GPT-2-medium \citep{radford2019language} and GPT-2-distilled \citep{sanh2019distilbert}. GPT-2-based models achieve strong results on a variety of natural language processing tasks such as question answering, summarization, and translation, without any specific training beyond next-word prediction.  
Furthermore, we analyze GPT-2-based models to allow for a direct comparison with prior brain alignment research \cite{goldstein2022shared, schrimpf2021neural}. We will present the averaged results across the three types of models in the main paper, and the individual results in the Appendix \ref{appendix:gpt2_small}, \ref{appendix:gpt2distilled}, \ref{appendix:gpt2medium}. We observed that as the model size increases, the effect of the stimulus-tuning perturbation as well as the residual effect after the contrast decrease. This reduction in effect is likely due to the small size of the dataset that we use for fine-tuning, which limits the learning capacity of larger models that already have a better next-word prediction ability. Therefore, we did not include additional larger models. For the baseline models we use the checkpoints provided by Huggingface \cite{wolf2020huggingfaces}\footnote{\url{https://huggingface.co/openai-community/gpt2}}\footnote{\url{https://huggingface.co/openai-community/gpt2-medium}}\footnote{\url{https://huggingface.co/distilbert/distilgpt2}}.

\subsection{fMRI data}\label{subsec:methods_data}
To evaluate the brain alignment of GPT-2 and of its perturbations, we use publicly available fMRI data provided by \citealp{wehbe2014simultaneously}, one of the largest publicly available fMRI datasets in terms of samples per participant. fMRI data were obtained from eight participants as they read chapter 9 of Harry Potter and the Sorcerer's Stone \citep{rowling1998harry} word-by-word. One fMRI image (TR) was acquired every 2 seconds (TR $= 2$ sec). The chapter was divided into four runs of approximately equal length and participants were allowed a short break at the end of each run. Each word of the chapter was presented for 0.5 seconds, after which a new word was presented immediately.

\subsection{Evaluation tasks}\label{subsec:methods_tasks}
We use two tasks to evaluate models: prediction of the next-word and brain alignment. Importantly, both tasks are evaluated using the same text, which corresponds to the fMRI stimulus. For consistency, we use the same setting to evaluate both next-word prediction and brain alignment: we evaluate each metric as described below using sliding windows of $20$ consecutive words (overlapping by 16 words, which corresponds to 4 TRs). We choose this window length because previous work has shown that using contexts larger than 20 words does not substantially improve brain prediction performance with similarly-sized language models \citep{toneva2019interpreting}. We empirically verified that this also holds in our setting.

\paragraph{Next-word prediction.}
To generate the next token, we follow best practices for GPT-2-based models which consist of a linear prediction head with weights tied to the input embeddings \citep{wolf-etal-2020-transformers}.  We evaluate the next-word prediction performance using the cross-entropy measure.

\paragraph{Brain alignment.}
To measure the brain alignment between a GPT-2-based model and the fMRI recordings, we employ a standard linear prediction head on top of the last transformer block. This prediction head learns a function that maps input stimulus representations to output brain recordings and is frequently used to measure how well word representations obtained from a language model can predict brain recordings \citep{jain2018incorporating,toneva2019interpreting,schrimpf2021neural}.
Similarly to previous work, we parameterize this function as a linear function, regularized using the ridge penalty. We train this function in a cross-validated way and test its performance on the data that was heldout during training. We select the ridge parameter via nested cross-validation. As a result, for each participant, we train four functions, then we aggregate the predictions and evaluate the brain alignment. We provide further details about this prediction head in Appendix \ref{appendix:prediction_head}.
We evaluate the brain alignment using Pearson correlation, computed between the predictions of heldout fMRI recordings and the corresponding true data. Specifically, for a model $q$ and voxel $v_j$ with corresponding heldout fMRI $y_j$, the brain alignment is computed as
\begin{equation*}
    \texttt{brain alignment}(q,v_j) = \texttt{corr}(\hat{y_j}, y_j),
\end{equation*}
where $\hat{y_j} = q(X)W_{q,j}$, $X$ is the input text sample to model $q$, and $W_{q,j}$ are the learned prediction weights corresponding to the voxel. All voxel-wise brain alignment scores are visualized on each participant's brain surface using PyCortex \citep{gao2015pycortex}.

\subsection{Perturbations}\label{subsec:methods_perturbations}
We aim to disentangle the effects on brain alignment of different types of information contained in language models, that we describe in Section \ref{sec:introduction}: next-word prediction, word-level information, and multi-word information. To achieve this, we designed two perturbations that isolate the contributions of these different types of information to the brain alignment when used as part of carefully designed contrasts (see Section \ref{subsec:contrasts_expected_effects}).

\paragraph{Input scrambling.}
The aim of the first perturbation is to control for the effect of word-level information on brain alignment. This perturbation consists of scrambling the words at inference time in each text sequence that we use to predict one fMRI TR image (i.e., 20 consecutive words). The order of words has been shown to be marginally important for other downstream tasks at inference time \citep{sinha-etal-2021-unnatural}. Therefore, if the words are scrambled, even though the next-word prediction ability will decrease, we expect the model to still predict the next word above chance level, using information from the 20-word context.

\paragraph{Stimulus-tuning.}
The second perturbation finetunes the baseline pretrained model with the next-word prediction objective on a training portion of the stimulus text. 
To perform the stimulus-tuning, we select training samples that consist of non-overlapping sequences of 80 consecutive words. For each baseline model, we trained four models, one for each held-out run of the fMRI data (see Appendix \ref{appendix:hyperparameters} for more details on model training). 

We expect stimulus-tuning to improve all three brain-relevant types of information we consider (next-word prediction, word-level information, and multi-word information). Therefore, we also expect that the stimulus-tuned model will also exhibit better brain
alignment than the baseline. However, stimulus-tuning itself is not sufficient to investigate the independent effect of either type of information on brain alignment. This perturbation is useful when combined with the input scrambling perturbation. By ensuring that the drop in next-word prediction accuracy after the scrambling perturbation is similar to the drop in the baseline after scrambling, we can control for both next-word prediction and word-level information (see Section \ref{subsec:contrasts_expected_effects}).

\subsection{Contrasts to disentangle brain alignment factors}\label{subsec:contrasts_expected_effects}
\input{contrasts}

\begin{figure*}[t]
  \centering
  \includegraphics[width=\textwidth]{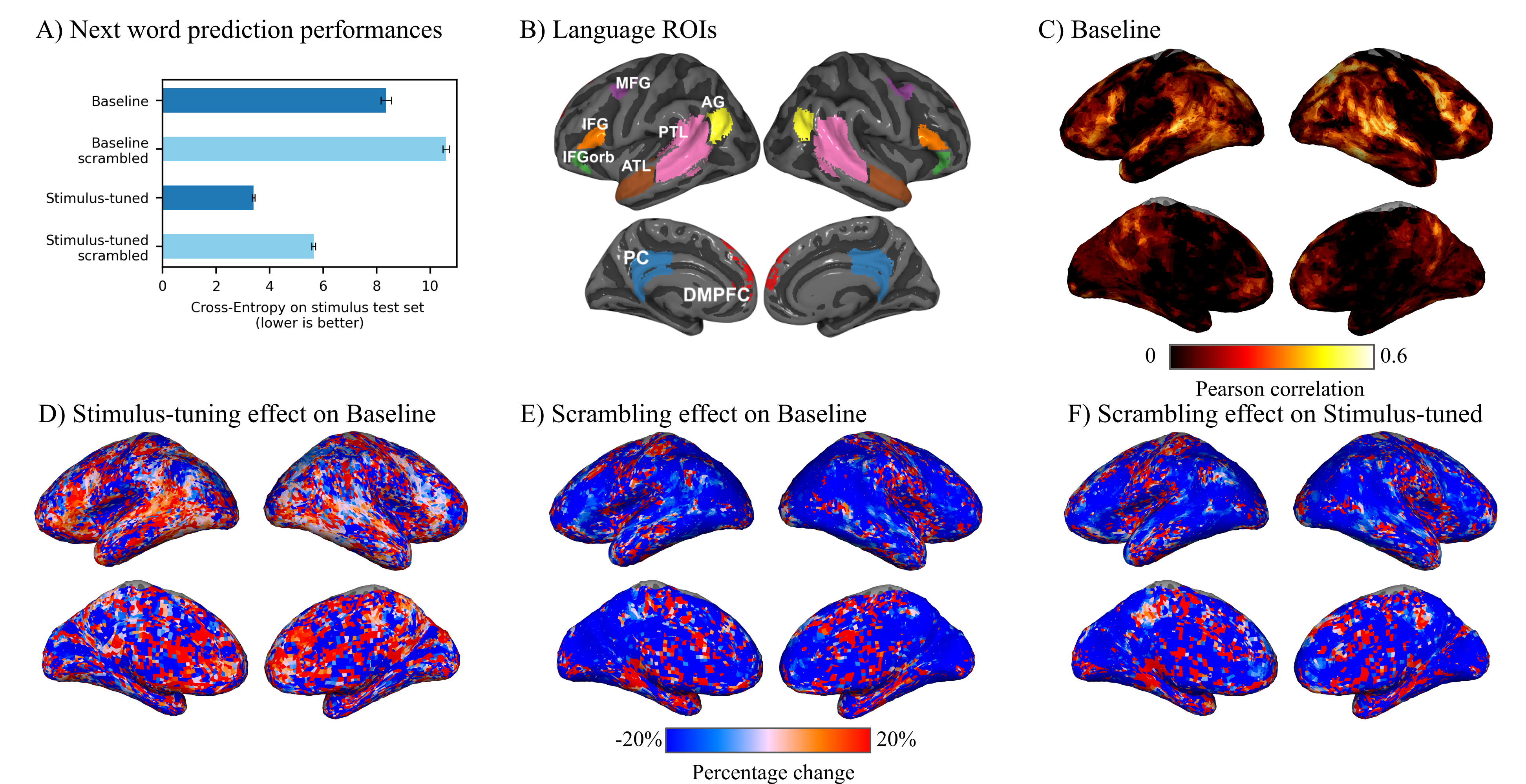}
  \caption{Performances of the GPT-2-small baseline and perturbed models at next-word prediction averaged across runs with standard deviation (A) and brain alignment (C-F). Stimulus-tuning improves both the next-word prediction (stimulus-tuned vs baseline in (A)) and brain alignment (D). Instead, scrambling reduces the next-word prediction (baseline vs baseline scrambled in (A)) and reduces the brain alignment (E and F). Despite the reduction in alignment due to the scrambling perturbation, all four models  exhibit alignment in language processing regions (B) (see Appendix Figure \ref{fig:appendix_all_subjects_gpt2_small} for brain alignment plots for all participants and Appendix Figures \ref{fig:appendix_all_subjects_gpt2_distill}, \ref{fig:appendix_all_subjects_gpt2_medium} for other models.}
  \label{fig:brainI_ROI_lm}
\end{figure*}
\section{Results}\label{sec:results}

In this section,  we report the results averaged across models. The results for the individual models can be found in Appendices \ref{appendix:gpt2_small}, \ref{appendix:gpt2distilled}, \ref{appendix:gpt2medium}.

\subsection{Next-word prediction}\label{subsec:next_word_prediction}
 
In Figure \ref{fig:brainI_ROI_lm}A, we report the next-word prediction performances of the GPT-2-small model and the corresponding perturbed models. The results for GPT-2-distilled and GPT-2-medium are consistent and reported in Appendix Figures \ref{fig:lm_gpt2_distill}, \ref{fig:lm_gpt2_medium}. We observe that the stimulus-tuned model performs better than the baseline. This verifies that stimulus-tuning indeed improves the model's ability to predict the next word in the heldout stimulus set. 

As expected, the performance of the scrambled models is worse than their unscrambled counterparts (i.e. baseline scrambled vs. baseline and stimulus-tuned scrambled vs. stimulus-tuned). We further observe that the next-word prediction performance of the stimulus-tuned scrambled model is still better than that of the baseline, indicating that the information gained by stimulus-tuning is not entirely counteracted by the scrambling perturbation.

\subsection{Brain alignment} \label{subsec:brain_alignment}
 
Figures \ref{fig:brainI_ROI_lm}C-F visualize the brain alignment  of the baseline model (i.e., Pearson correlation between predicted and true brain recordings) and the percentage change between pairs of models for one participant for GPT-2-small. Results for the remaining participants and models are largely consistent and are shown in Appendix Figs. \ref{fig:appendix_all_subjects_gpt2_small}, \ref{fig:appendix_all_subjects_gpt2_medium}, \ref{fig:appendix_all_subjects_gpt2_distill}. 

\begin{figure}[t]
\centering
\includegraphics[width=0.5\textwidth]{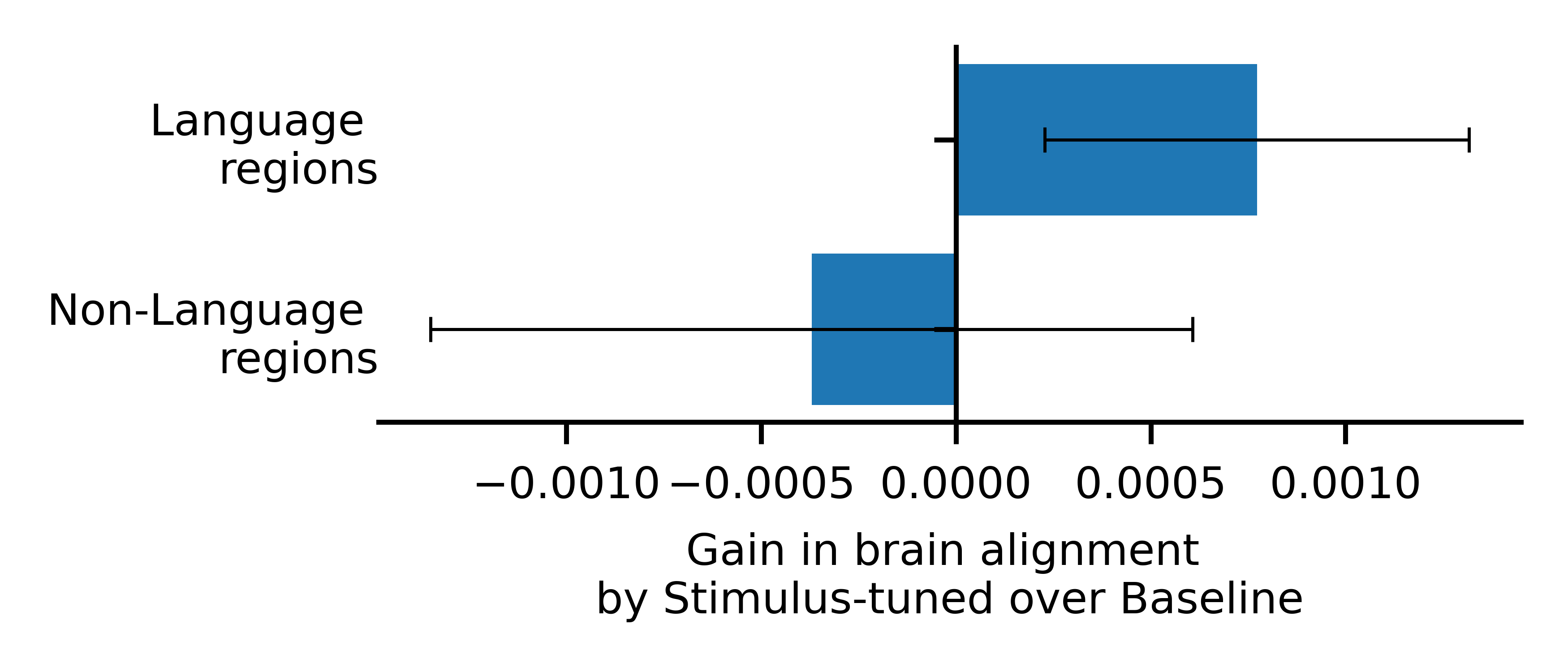}
\caption{Impact of the stimulus-tuning perturbation on the baseline model. For each model (GPT-2-small, medium, distill) we computed the median difference in language and non-language regions across participants. Here we display the average difference across models as well as the standard deviation. Results for the single models are reported in Appendix Figures \ref{fig:whole_text_vs_base_gpt2_small}, \ref{fig:whole_text_vs_base_gpt2_distill}, \ref{fig:whole_text_vs_base_gpt2_medium}.}
  \label{fig:whole_text_vs_base_gpt2}
\end{figure}

\begin{figure}[t]
\centering
\includegraphics[width=0.5\textwidth]{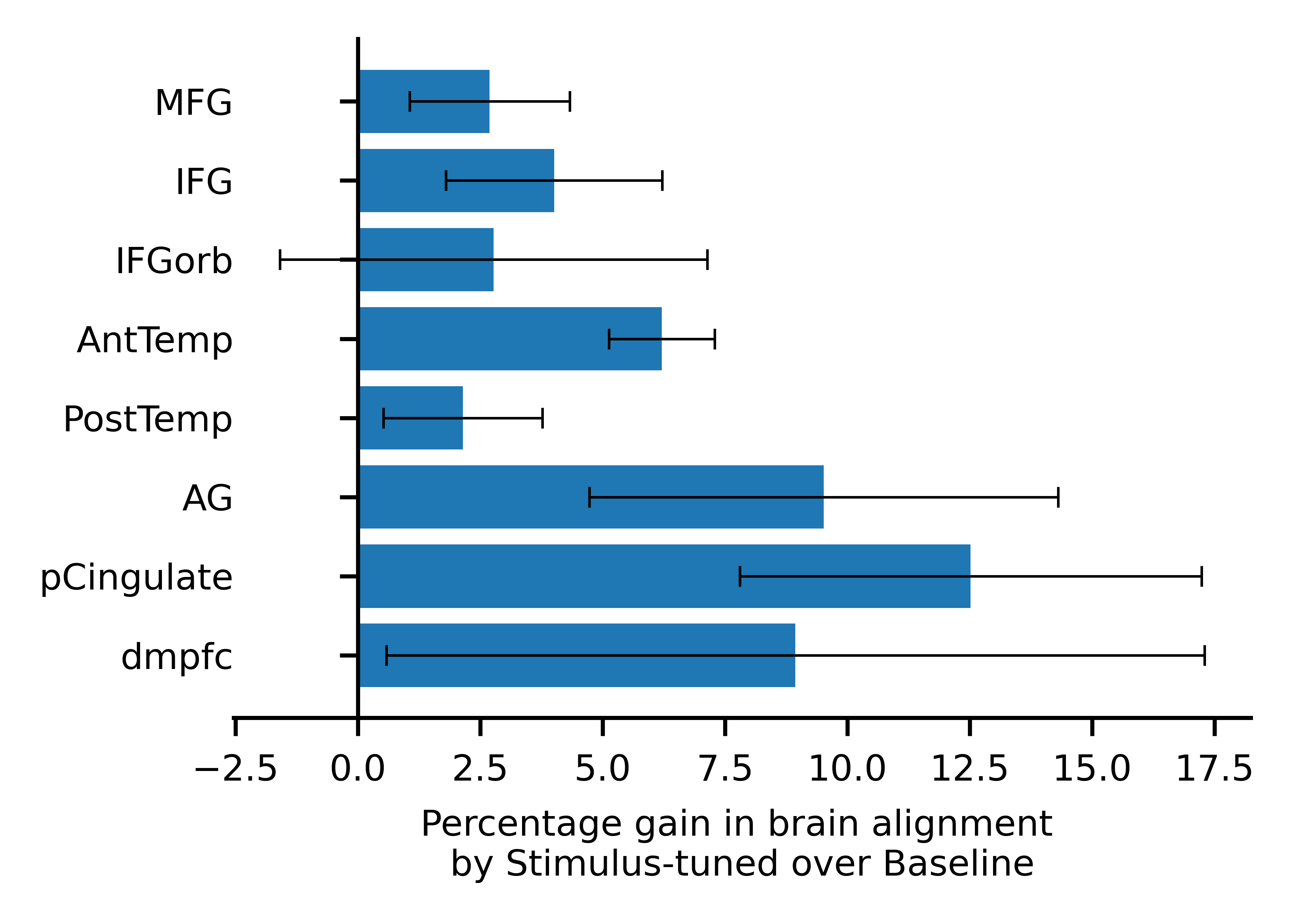}
\caption{Impact of the stimulus-tuning perturbation on the baseline model. For each model (GPT-2-small, medium, distill) we computed the median percentage gain by stimulus-tuned over baseline in language regions across participants. Here we display the average percentage gain across models as well as the standard deviation. We include only voxels with estimated noise ceiling values >0.05. Results for the single models are reported in Appendix Figures \ref{fig:roi_text_vs_base_gpt2_small},  \ref{fig:roi_text_vs_base_gpt2_distill}, \ref{fig:roi_text_vs_base_gpt2_medium}.}
  \label{fig:roi_text_vs_base_gpt2}
\end{figure}

\paragraph{Effects of stimulus-tuning.} In Figure \ref{fig:brainI_ROI_lm}D, we observe that the stimulus-tuned model aligns better with the brain recordings than the baseline, particularly in many brain areas that have been previously implicated in language-specific processing \citep{fedorenko2010new,fedorenko2014reworking} and word semantics \citep{binder2009semantic} (visualized in Figure \ref{fig:brainI_ROI_lm}B and listed in Appendix \ref{appendix:list_roi}). We quantify the improvement across models and participants in brain alignment due to stimulus-tuning in language processing regions versus non-language processing regions in Figure \ref{fig:whole_text_vs_base_gpt2}. 
Here, we demonstrate that the stimulus-tuned model has higher brain alignment in language-related regions than in non-language-related regions. This indicates that the stimulus tuning perturbation contributes to an improvement in the model's performance, particularly in language-related ROIs. The results shown are computed using all voxels in the brain, including a large number of noisy voxels. Therefore, the reported numbers are numerically low. To focus on more informative voxels, we quantify the difference in each language-related ROI and use an estimate of the noise ceiling in each voxel to discard noisy voxels (see Appendix \ref{appendix:noise_ceiling} for details). In Figure \ref{fig:roi_text_vs_base_gpt2}, we present the average percentage gain in brain alignment due to stimulus-tuning across models in each specific language ROI. Here we include only voxels that have estimated noise ceiling values > 0.05. The figure reveals a positive gain of the stimulus-tuned model over the baseline in every language region.

The results show that stimulus-tuning leads to both an improved ability to predict the next word and an improved alignment with fMRI recordings, but we are not able to conclude that the improvement in alignment with the brain is due to the improved prediction of the next word. The reason is that improving a model's ability to predict the next word may also improve other aspects of the model that are brain-relevant, such as its ability to represent word-level or multi-word information, that are specific to the stimulus narrative.

\paragraph{Effects of scrambling.} In Figure \ref{fig:brainI_ROI_lm}E-F, we observe that for both the baseline and stimulus-tuned model the scrambling perturbation affects the alignment, particularly in the language-related ROIs (see Figure \ref{fig:brainI_ROI_lm}B).

However, both the baseline scrambled and stimulus-tuned scrambled models still align with the fMRI recordings, particularly in the language ROI, as shown in Appendix Figures \ref{fig:appendix_all_subjects_gpt2_small}, \ref{fig:appendix_all_subjects_gpt2_distill}, \ref{fig:appendix_all_subjects_gpt2_medium}. This suggests that even when perturbing the next-word prediction capability and multi-word information, a language model is able to strongly align with brain areas that are thought to process language.

Interestingly, we observe that the effect of scrambling on the stimulus-tuned model is much larger for brain alignment than for next-word prediction. For next-word prediction, the stimulus-tuned scrambled performs worse than the stimulus-tuned model but better than the baseline. In Appendix Figures \ref{fig:appendix_all_subjects_gpt2_small}, \ref{fig:appendix_all_subjects_gpt2_distill}, \ref{fig:appendix_all_subjects_gpt2_medium}, we see that the stimulus-tuned scrambled performs worse at brain alignment, not only with respect to the stimulus-tuned model but also to the baseline.
This is an initial hint that next-word prediction is not the only key information in aligning language models and brain recordings.

We show that scrambling affects both the next-word prediction ability and the brain alignment of language models. However, we are not able to draw a conclusive link yet. The reason is that the scrambling procedure only controls for word-level information, but not for any possible changes in multi-word information, which may also contribute to the decrease of alignment with the language processing brain areas.

\subsection{Controlling for both word-level information and next-word prediction}

\begin{figure}[t]
\centering
\includegraphics[width=0.5\textwidth]{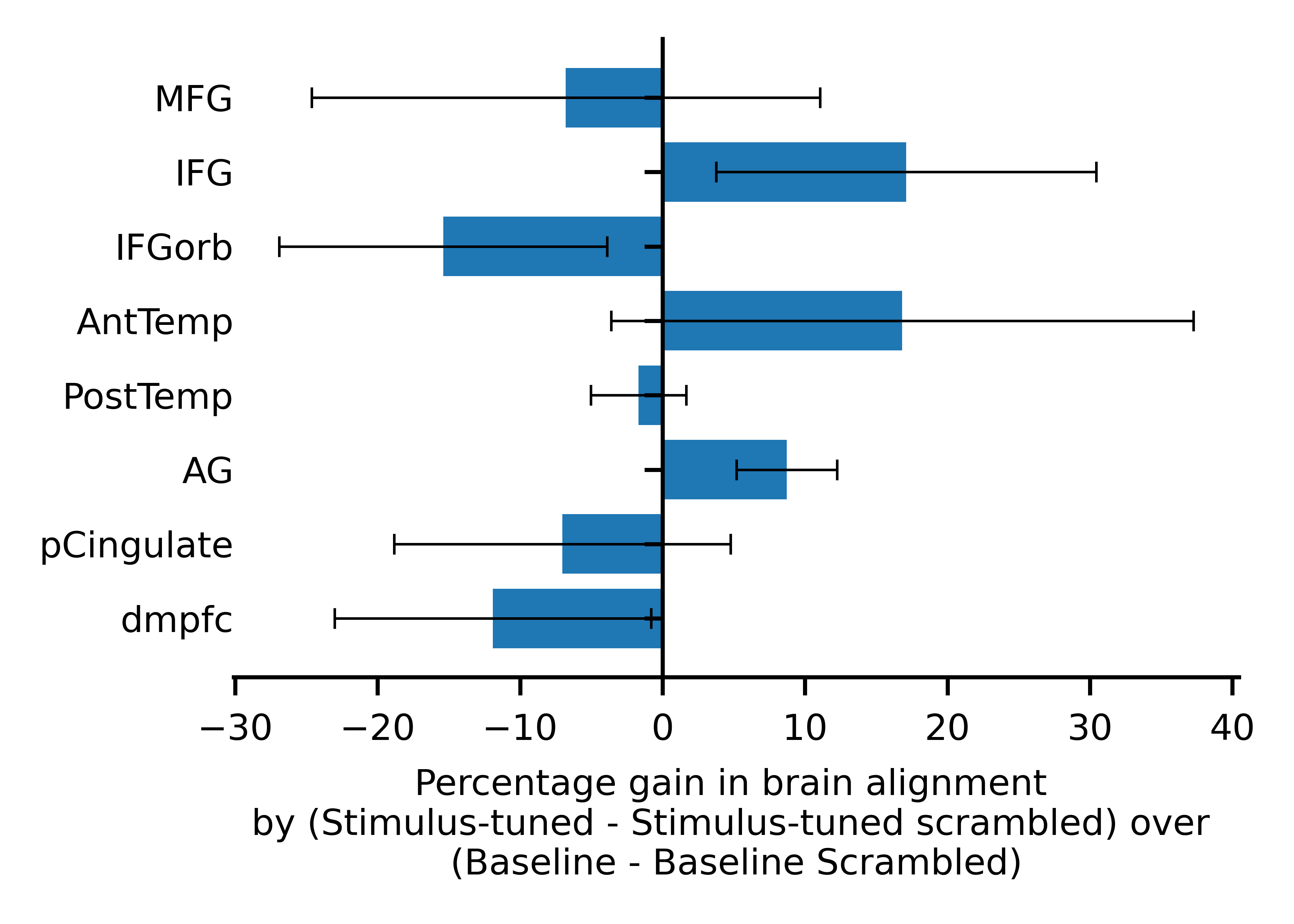}
\caption{Impact of the scrambling perturbation on the stimulus-tuned model versus its impact on the baseline model. For each model (GPT-2-small, medium, distill) we computed the median percentage gain by (stimulus-tuned - stimulus-tuned scrambled) over (baseline - baseline scrambled) in language regions across participants. Here we display the average percentage gain across models,  as well as the standard deviation. We include only voxels with estimated noise ceiling values >0.05. Results for the single models are reported in Appendix Figures \ref{fig:roi_final_gpt2_small},  \ref{fig:roi_final_gpt2_distill}, \ref{fig:roi_final_gpt2_medium}.}
  \label{fig:roi_final_gpt2}
\end{figure}

In Figure \ref{fig:roi_final_gpt2}, we report the average percentage gain by (stimulus-tuned - stimulus-tuned scrambled) over (baseline - baseline scrambled) ($\Delta_{}^{stim}$ - $\Delta_{}^{base}$), across GPT-2 models, in each specific ROI, including voxels that have estimated noise ceiling values > 0.05 (see Figures \ref{fig:appendix_final_contrast_gpt2_small}, \ref{fig:final_contrast_gpt2_distill}, \ref{fig:final_contrast_gpt2_medium} in Appendix for the corresponding brain plot of each model). Given the high variability across subjects and the number of subjects in our setting, obtaining statistically significant results is challenging (see Appendix \ref{app:stat} for details). Despite these challenges we observe that there is still a positive residual brain alignment after the contrast for two language processing regions, the Inferior Frontal Gyrus (IFG) and Angular Gyrus (AG), even when controlling for next-word prediction and word-level information, across three different models. This is evidence that the alignment with the language model in these areas is due to more than next-word prediction and word-level information (See Eq. \ref{eq:final}). 

\section{Related Works} \label{sec:related_works}
Several previous studies have investigated the alignment between pretrained language models and brain recordings of people comprehending language, finding significant similarities 
\citep{wehbe2014aligning,jain2018incorporating,toneva2019interpreting,abdou2021does,schrimpf2021neural,hosseini24}. Our work builds on these and further studies the reasons for these similarities. The work of \citet{goldstein2022shared} is most directly related to our research question, as they suggest that the prediction of the next word is a key reason for the alignment between language models and brain recordings, based on evidence that ECoG electrodes can predict the neural network representation of upcoming words in a story. Our work uses perturbations to disentangle next-word prediction from other types of information that may affect brain alignment (word-level and multi-word information) and offers an additional account of the necessary information for brain alignment.

Our work also relates to a growing body of research on disentangling the contributions of different types of information to the alignment between brain recordings and language models. \citet{toneva2020combining} present an approach to disentangle supra-word meaning from lexical meaning showing that the supra-word meaning is predictive of fMRI recordings in two language regions (Anterior and Posterior Temporal Lobes), which was further adapted by \citet{oota2023speech} and \citet{oota2024joint} to disentangling effects of other linguistic properties. \citet{caucheteux2021decomposing} and \citet{reddy2021can} aim to disentangle alignment due to syntactic and semantic processing. \citet{toneva2022same} examine whether representations obtained from a language model align with different language processing regions in similar or different ways. \citet{Kauf2023.05.05.539646} investigate the contribution of word-level semantics to the brain alignment of language models, showing that syntactic perturbations have a lesser impact on brain alignment compared to semantic perturbations. \citet{gauthier-levy-2019-linking} demonstrate that fine-tuning language models on scrambled data has been shown to be beneficial for brain decoding. Our experiments reveal that the scrambling perturbation influences brain alignment. Despite this, when sentences are scrambled at inference time, the model retains the capability to predict brain responses in regions associated with language processing. However, the aim of our work is not to directly evaluate the significance of semantics or syntax, but rather to investigate the effects on brain alignment when information relevant to next word prediction is controlled for. Therefore, our proposed perturbations are complementary to these previous works and may yield additional insights if combined.

Other studies have used perturbations related to word order to investigate some properties of language models. \citet{pandia2021sorting} introduced distracting content to test how robustly language models retain and use that information for prediction, showing that language models are particularly susceptible to semantic similarity and word position. \citet{papadimitriou2022classifying} applied a perturbation (scrambling method) to investigate where the semantic and syntactic processing is taking place in BERT, revealing that early layers care more about the lexicon, while the latter layer care more about word order.
Our current work contributes to this research direction by examining the effects of scrambling on both brain alignment and language modeling.

Finally, a work by \citet{aw2023summary} finetunes language models to summarize narratives and finds improved brain alignment, despite a lack of improvement in next-word prediction. While this finding suggests a similar conclusion to the one from our work--that next-word prediction performance is not necessary for improved brain alignment--the perturbation approach in our work allows additional control over the language model representations and is complementary to this previous work.

\section{Discussion} \label{sec:discussion}
We showed that the perturbation that we termed \emph{stimulus-tuning} (i.e., finetuning a pretrained model on a validation stimulus text) can increase the alignment with brain recordings that correspond to a heldout text, particularly in several language processing brain areas. We quantified this improvement by comparing the stimulus-tuned model and the baseline in these brain areas. Stimulus-tuning may improve brain alignment due to improved ability to represent the next word, previously seen individual words, or multi-word information that are specific to the stimulus narrative. 

Using the perturbation that we termed \emph{input scrambling}, we showed that the improved next-word prediction capabilities of the stimulus-tuned model is not the only reason for improved brain alignment. 
We showed that leveraging a contrast that controls for word-level information and next-word prediction, we still obtain a residual brain alignment.
Specifically, we show that, across multiple GPT-2 models, improvements in alignment with brain recordings in two language regions—Inferior Frontal Gyrus (IFG), Angular Gyrus (AG)—(see Figure \ref{fig:roi_final_gpt2}) are due to more than improvements in next-word prediction and word-level information. 

One possible reason for this improvement in brain alignment is improved capabilities to represent multi-word information that are specific to the stimulus text. This hypothesis aligns with previous work that has found the Inferior Frontal Gyrus (IFG) to be sensitive to syntax \citep{friederici2003role,friederici2012cortical} and the Angular Gyrus (AG) to multi-word event structure \citep{ramanan2018rethinking,humphreys2021unifying}. Note that the fact that we do not find strong effects in other language regions does not necessarily mean that they do not process multi-word information.

\section{Conclusion}
This work aims to deepen our understanding of the existing alignment between language models and brain recordings. We proposed two perturbations to pretrained language models that, when used together, can control for the effects of next-word prediction and word-level information on the alignment with brain recordings. Using these controls, we show that improvements in brain alignment are due to more than improvements in next-word prediction and word-level information. Our findings are relevant for both cognitive neuroscientists as well as natural language processing researchers.

The findings are relevant for cognitive neuroscientists because they suggest that accurate prediction of the next word is not a necessary condition for brain alignment. It is possible that learning to accurately predict the next word is sufficient for inducing other properties in the language model representations that are necessary for brain alignment, such as multi-word information, and future work can further examine this hypothesis.

Our findings are also relevant to NLP researchers who examine what language models can learn from only text. We show that finetuning a language model with small amounts of text can increase its alignment with never-before-seen brain recordings, and that this improvement in brain alignment is not purely due to next-word prediction or word-level information. This finding suggests that training a language model with little additional text can improve its representations of multi-word information in a brain-relevant way. We note that while our methodology controls for next-word prediction ability at inference time, it still relies on the next-word prediction objective during training. However future work can investigate alternative training objectives, that may improve their ability to represent multi-word information in other ways. One example is work by \citet{aw2023summary} that shows that finetuning a language model using a summarization objective can further improve brain alignment.

\section{Limitations}\label{sec:limitations}
We have attempted to address potential limitations in our research design, however, it is important to acknowledge the inherent limitations of our study. Firstly, we use GPT-2 based models to compare with previous work using the same model family. 
However, analyzing additional language models, such as larger language models, or ones trained with a masked language modeling objective, is an important next step for insights that can be generalized to larger families of models, even if previous work has suggested that larger language models could diverge from human-like representations \citep{oh2023transformerbased}. Moreover, in our highly controlled setting, stimulus-tuning larger language models could lead to a smaller increase in next-word prediction and brain alignment, given the relatively small amount of data available for finetuning. This is visible when comparing the stimulus-tuning improvements of GPT-2-distilled and GPT-2-small with GPT-2-medium: as the model size increases, the effect of stimulus tuning and the residual effect after the contrast decreases. 

Secondly, our experiments were conducted with one fMRI dataset. Even though the dataset we chose is a well studied dataset, is one of the largest ones available, and care was taken to test the generalization performance to never-before-seen brain data, the effects we observe may still be specific to this dataset. Testing datasets that differ in text genre (we use only a narrative dataset) and language (our conclusions are drawn for English text only) would be particularly interesting. 

Thirdly, our findings are based on some experimental choices, such as the scrambling method. For instance, \citet{mollica2020} showed that fMRI activity in humans reading scrambled sentences remains relatively stable under certain perturbations. Therefore, further investigation into different scrambling methods and their effects could provide additional insights. Furthermore, despite the presence of a strong positive correlation between next-word prediction and brain alignment reported by \citet{schrimpf2021neural} and our experiments (0.61 Pearson correlation, see Appendix \ref{appendix:linear_relation}), this relationship is not perfectly linear so it is possible that the subtraction that we employ does not perfectly control for the effect of the next-word prediction capabilities.

Fourthly, our results are based on observing changes in NWP on a held-out stimulus test set. However, changes in next-word prediction can be influenced by multiple factors. The model could acquire general English knowledge or knowledge specific to the Harry Potter chapter. Since we are using a language model heavily pretrained on general English knowledge, we believe that the improvement in next-word prediction ability is primarily due to learning about the specific domain we are fine-tuning for. However, investigating further the causes behind the increase in next-word prediction performance could be an insightful next step.

\section{Ethics and Broader Impact}
Our research impact is closely related to its potential social implications. We propose a method to analyze language models, with the aim of gaining a better understanding of the differences and similarities between the human brain and neural networks. This understanding can serve two key purposes: firstly, shedding light on the reasons behind the impressive power of neural networks; secondly, enhancing our comprehension of the underlying mechanisms governing brain functions. A deeper understanding of both artificial neural networks and biological neural networks can significantly benefit society, especially considering the prevalence of black-box artificial intelligence systems. By leveraging insights from the human brain, we can strive to integrate these systems more consciously into human activities. This integration is essential for ensuring transparency, interpretability, and ethical use of AI, thereby fostering a positive and responsible impact on society.

\section*{Acknowledgements}
The authors would like to thank Shailee Jain and Sebastian Schuster for helpful feedback on an earlier version of this manuscript. Work by Gabriele Merlin was supported by the CS@max planck graduate center.
\bibliography{arxiv_acl_latex}

\appendix
\newpage
\onecolumn
\section{Prediction head}\label{appendix:prediction_head}

Similarly to previous work, to predict the fMRI recordings corresponding to a given TR, we use a prediction head that maps from the model representation to the fMRI space. We parameterize this function as a linear function, regularized using the ridge penalty. We train this function in a cross-validated way and test its performance on the data that was heldout during training. We select the ridge parameter via nested cross-validation. For each participant, we train four functions, each one using three of the four runs in the fMRI recordings, and reserve the remaining run for testing.
To generate the models representation we average the embeddings corresponding to each fMRI image (i.e., TR) and uses a concatenation of the previous $5$ averaged TR embeddings. The averaging is done in order to down-sample the word embeddings (words presented at 0.5 seconds) to the TR rate (2 seconds). The features of the words presented in the previous TRs are included to account for the lag in the hemodynamic response that fMRI records. Because the response measured by fMRI is an indirect consequence of brain activity that peaks about 6 seconds after stimulus onset, predictive methods commonly include preceding time points \citep{nishimoto2011reconstructing,wehbe2014simultaneously,huth2016natural}. This allows for a data-driven estimation of the hemodynamic response functions (HRFs) for each voxel, which is preferable to assuming one because different voxels may exhibit different HRFs.

\section{Training hyperparameters}
\label{appendix:hyperparameters}

To perform the stimulus-tuning, we select training samples that consist of non-overlapping sequences of 80 consecutive words. We train the models with a batch size of 32 and for 2 epochs and we saved checkpoints of the models for each batch. The stimulus text is divided into 4 consecutive sections to enable cross-validation. For each GPT-2-based models we stimulus-tuned four models, one for each held-out run. We train the models using the default hyperparameters provided by Huggingface. As mentioned in Section \ref{subsec:contrasts_expected_effects} we selected the checkpoint for the stimulus-tuned model that best satisfies $\delta^{stim}\approx \delta^{base}$.

\section{List of ROI related to language processing and word semantics}
\label{appendix:list_roi}
These regions include: Middle Frontal Gyrus (MFG), Inferior Frontal Gyrus (IFG), Inferior Frontal Gyrus par Orbitalis (IFGorb), Anterior Temporal Lobe (AntTemp), Posterior Temporal Lobe (PostTemp), Angural Gyrus (AG), Posterior Cingulate Cortex (pCingulate), Dorsomedial Prefrontal Cortex (dmPFC) (see Figure \ref{fig:brainI_ROI_lm}B).

\section{Significance testing and Participants variability}\label{app:stat}

We designed the experiments to compare different models and their perturbations, testing their capabilities in brain alignment and displaying the percentage gain of one model over another. For each comparison, we conducted significance testing using ROI-level Wilcoxon signed-rank tests with p < 0.05 and Holm-Bonferroni correction \cite{holm_79}. For the final contrast the significance test revealed statistically significant residual alignment in the IFG (pvalue of 0.027). After correcting for multiple comparisons across ROI, this pvalue was no longer statistically significant at a threshold of 0.05. Similarly for the baseline vs stimulus-tuned contrast the results are not significative.
Given the high variability across subjects and the number of subjects in our setting, obtaining statistically significant results is challenging.  Moreover, due to our controlled experimental design, we only finetuned the baseline model using the stimulus text, that is composed by few samples, which could result in a minimal effect from this perturbation. Still, positive residual alignment in the IFG and AG is observed across three different models: GPT-2-distill, GPT-2-small, and GPT-2-medium. Despite these challenges, we believe that our results are informative and are strengthened by the analysis across three different models.

\section{Linear relation brain alignment and cross-entropy loss}
\label{appendix:linear_relation}
\begin{figure}[ht]
    \centering
    \includegraphics[width=0.5\textwidth]{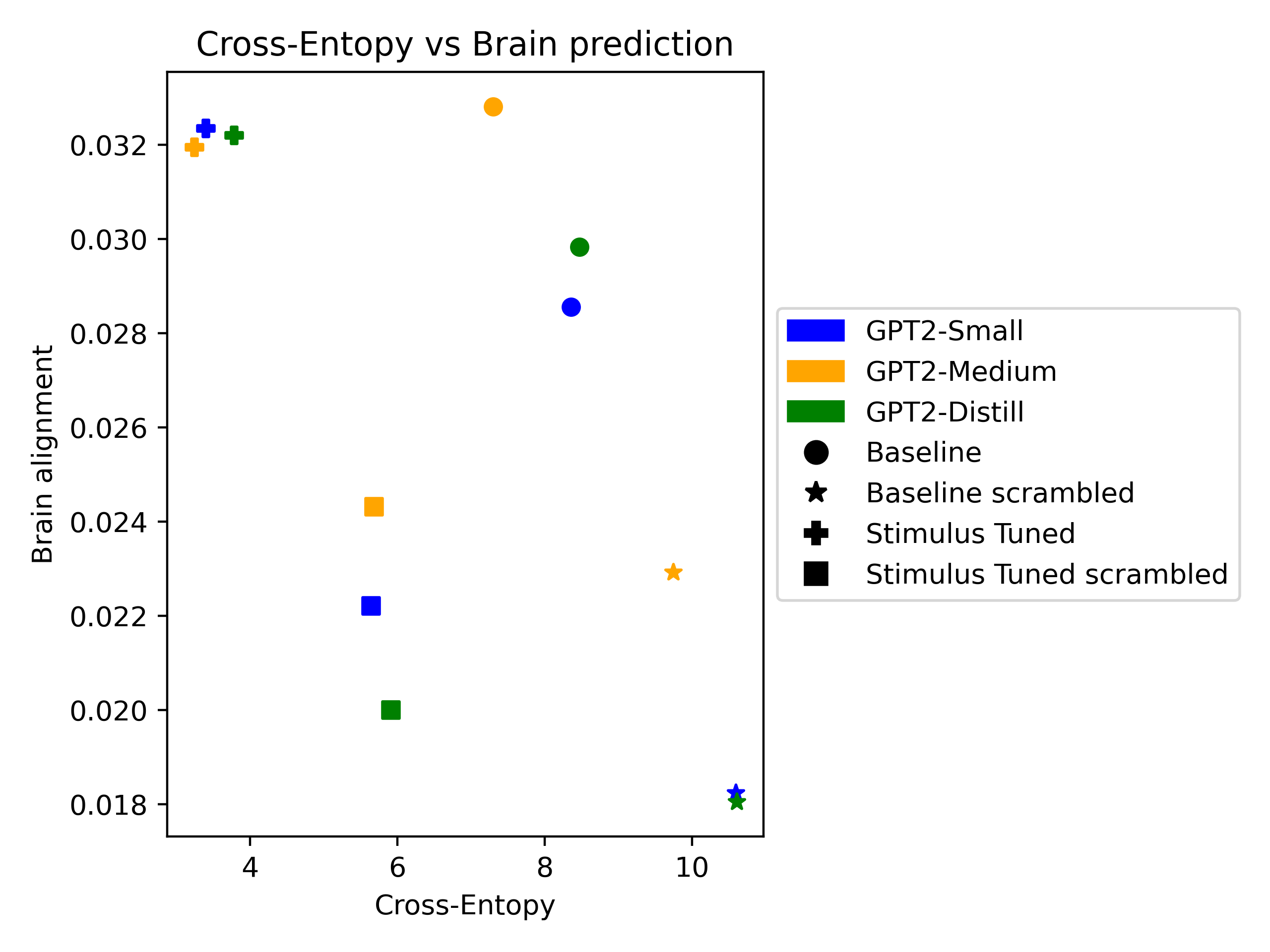}
     \caption{Correlation between the brain predictions capability of GPT-2-small, GPT-2-medium and GPT-2-distilled (on a held-out test set) and their cross-entropy loss. The correlation between these two measures is 0.61, similar to the correlation between the   brain predictions capability and the perplexity 0.67.}
    \label{fig:additional}
\end{figure}

\citet{schrimpf2021neural} shows a linear relationship between brain alignment and the next-word prediction capability of language models. Specifically, to test this relationship, they used Pearson correlation (normalized using estimated noise ceiling) for brain alignment and the exponentiated cross-entropy, i.e., perplexity, to evaluate next-word prediction capability. However, language models are typically fine-tuned using cross-entropy loss, and our experiments during fine-tuning showed that it is not feasible to achieve a similar difference in perplexity between (stimulus-tuned and stimulus-tuned scrambled) and (baseline and baseline scrambled) models. Therefore, in our final contrast, we employed the difference in cross-entropy between models. Although \citet{schrimpf2021neural} suggested a linear relationship between perplexity and brain alignment, this does not guarantee a linear relationship between cross-entropy and brain alignment, even if the two metrics are related. For this reason, we tested whether the linear relationship also holds for cross-entropy loss. We demonstrate in Figure \ref{fig:additional} that there is indeed a linear relationship between cross-entropy and brain alignment (0.61), albeit slightly lower than the one between brain alignment and perplexity (0.67). More experiments with additional models and datasets are necessary to verify if this relation holds in general, but for our analysis, it is sufficient to apply the subtraction explained in Section \ref{subsec:contrasts_expected_effects}.

\section{Cross perturbation contrast illustration}
\begin{figure}[H]
    \centering
    \includegraphics[width=\textwidth]{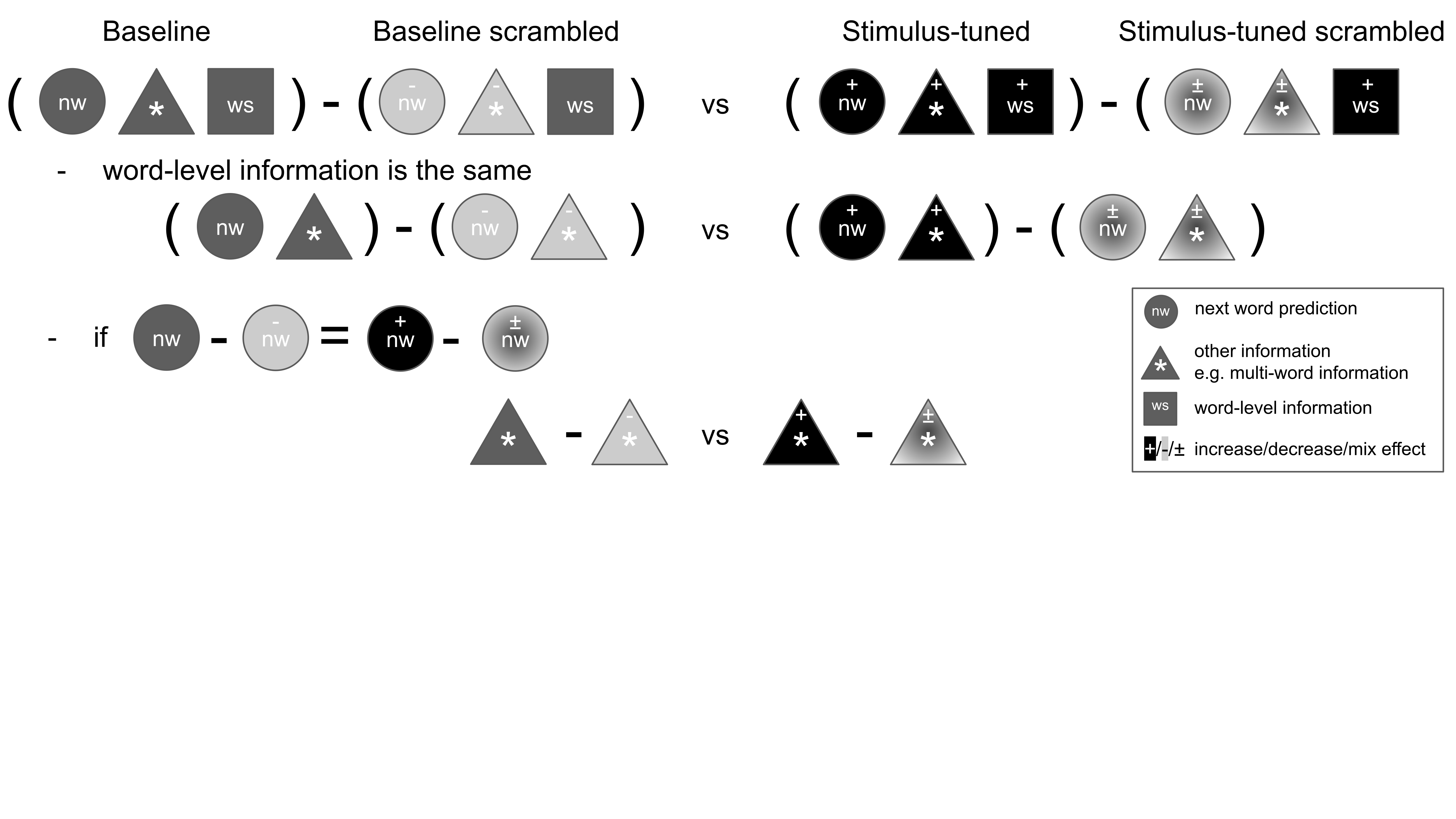}
    \caption{Illustration of the contrast (baseline - baseline scrambled) vs (stimulus-tuned - stimulus-tuned scrambled). Any observed effect of this contrast is controlled for word-level information information. Additionally, if the next-word prediction differences are equal then the contrast control also for next-word prediction. Therefore any observed effect would then be due to more than next-word prediction and word-level information information. An example of other source of information is the multi-word information.}
    \label{fig:cross_contrast_illustration}
\end{figure}

\section{Noise Ceiling estimation}\label{appendix:noise_ceiling}
The noise ceiling estimation is employed to assess the signal quality of fMRI data. fMRI data are inherently noisy, and the noise ceiling estimation provides an estimate of the variance that can be explained by an ideal data-generating model. The method relies on predicting the fMRI activity of a target participant using linear models trained on data from another participant. Linear encoding models are utilized. For a more detailed explanation, refer to \cite{schrimpf2021neural}. We employed this approach because our method relies on contrasts between different models aimed at predicting the brain activity of each subject, necessitating a consistent set of voxels for each encoding model.

\begin{figure}[ht]
    \centering
    \includegraphics[width=\textwidth]{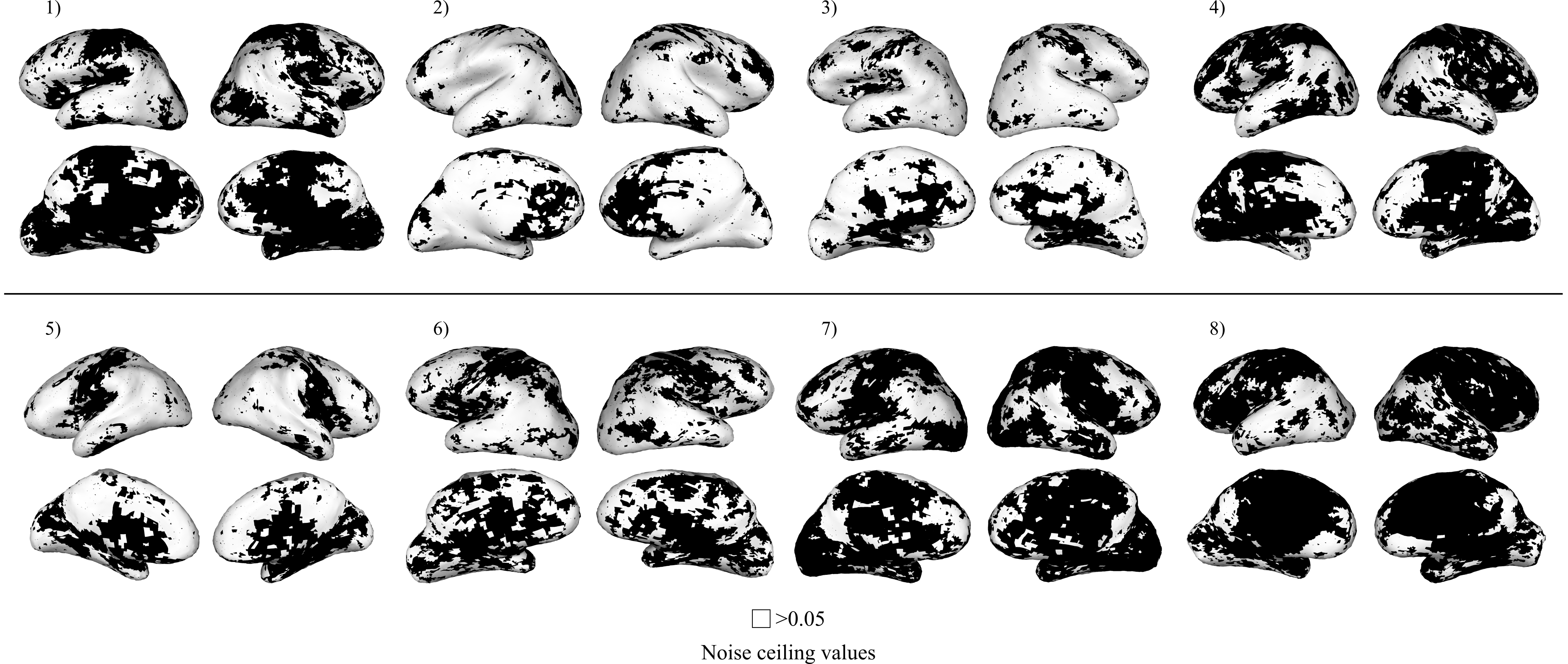}
    \caption{Voxel-wise estimated noise ceiling values. To exclude noisy voxels, we selected, for each participant, those with noise ceiling estimates above 0.05.}
    \label{fig:noise_celing}
\end{figure}

\section{GPT-2-small}\label{appendix:gpt2_small}
\begin{figure}[H]
    \centering
    \includegraphics[width=0.9\textwidth]{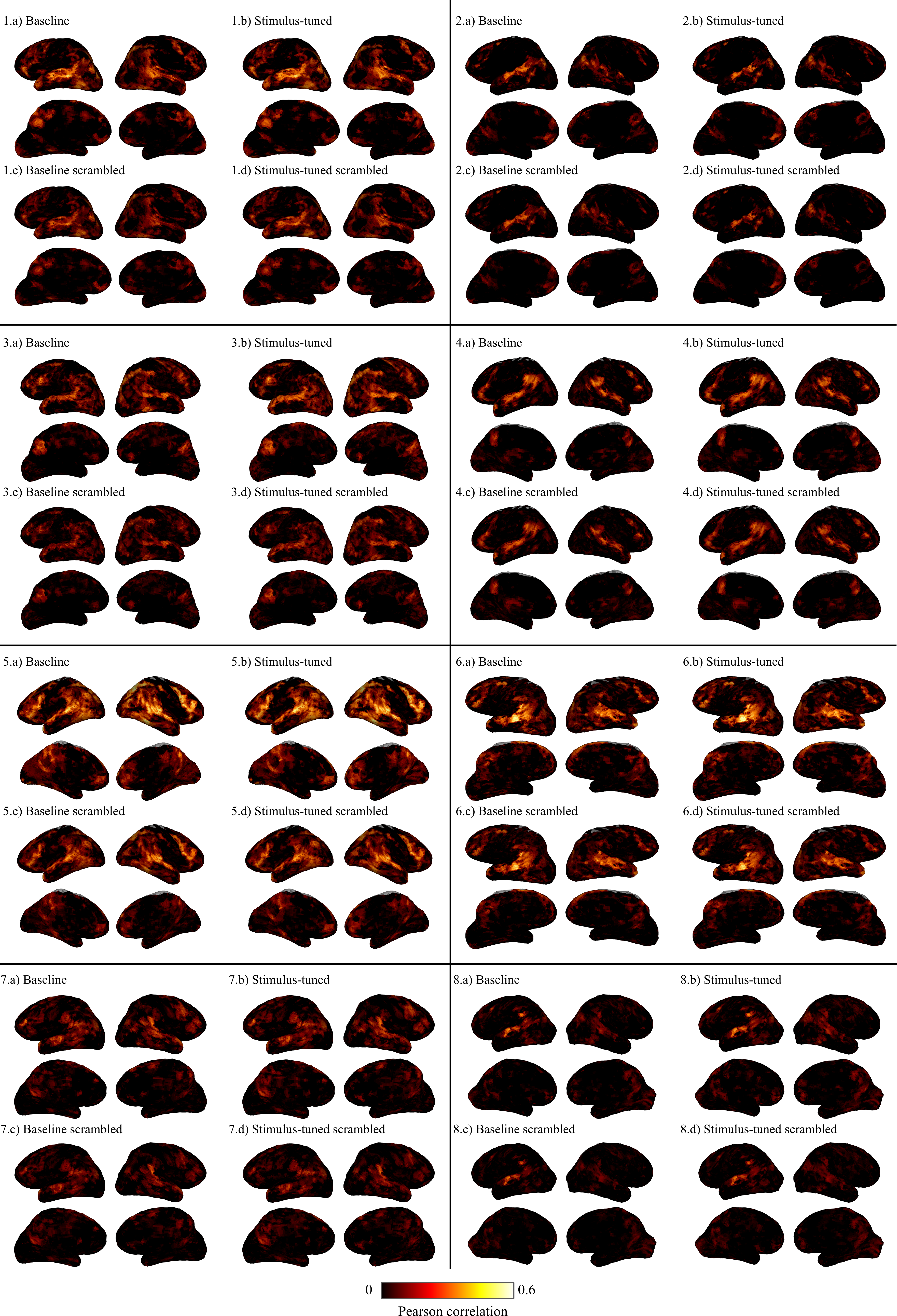}
  \caption{Performances of the GPT-2-small baseline and perturbed models of all participants at the brain alignment task. Stimulus-tuning improves the brain alignment (stimulus-tuned in (.b) vs baseline in (.a)) for almost all participants. In contrast, scrambling reduces the brain alignment (baseline in (.a) vs baseline scrambled in (.c)). Despite the reduction in alignment due to the scrambling perturbation, all four models (.a,.b,.c,.d) exhibit alignment in language processing regions.}
  \label{fig:appendix_all_subjects_gpt2_small}
\end{figure}

\begin{figure}[H]
\centering
\includegraphics[width=0.5\textwidth]{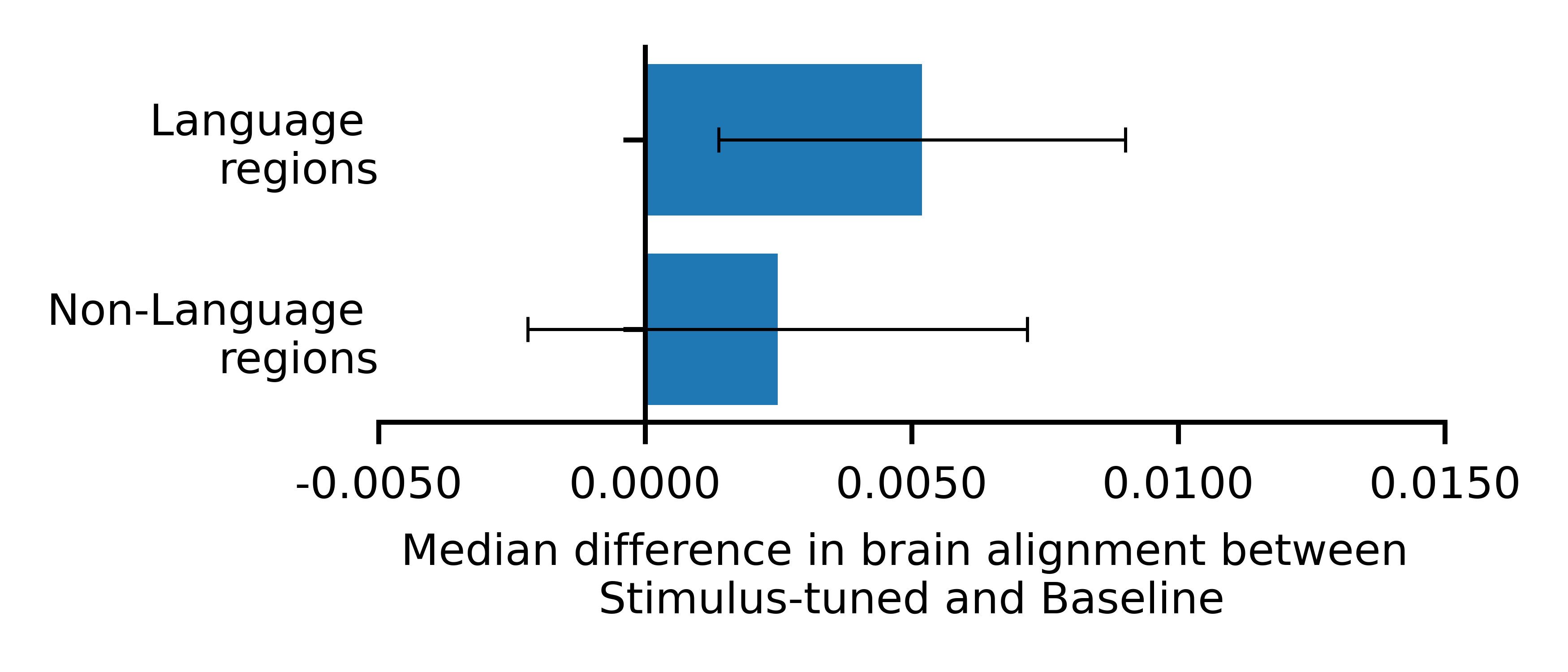}
\caption{Median difference in brain alignment between GPT-2-small stimulus-tuned and baseline models. We display the median difference in language and non-language regions and the median absolute deviation across the 8 participants.}
  \label{fig:whole_text_vs_base_gpt2_small}
\end{figure}

\begin{figure}[H]
\centering
\includegraphics[width=0.5\textwidth]{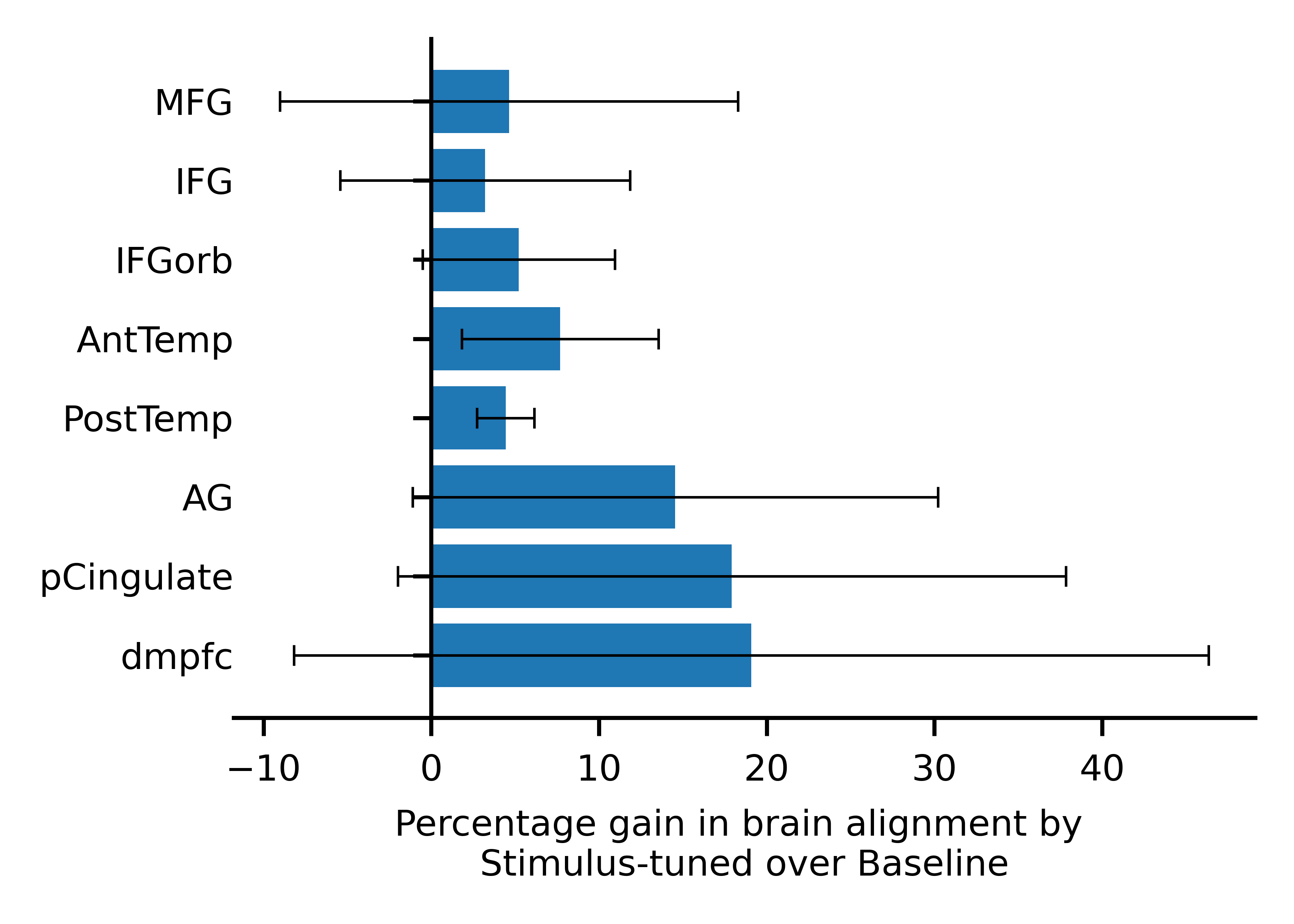}
\caption{Impact of the stimulus-tuning perturbation on the brain alignment of the GPT-2-small baseline model. We show the median percentage gain as well as the median absolute deviation across participants. We include only voxels with estimated noise ceiling values >0.05.}
  \label{fig:roi_text_vs_base_gpt2_small}
\end{figure}

\begin{figure}[H]
\centering
\includegraphics[width=0.5\textwidth]{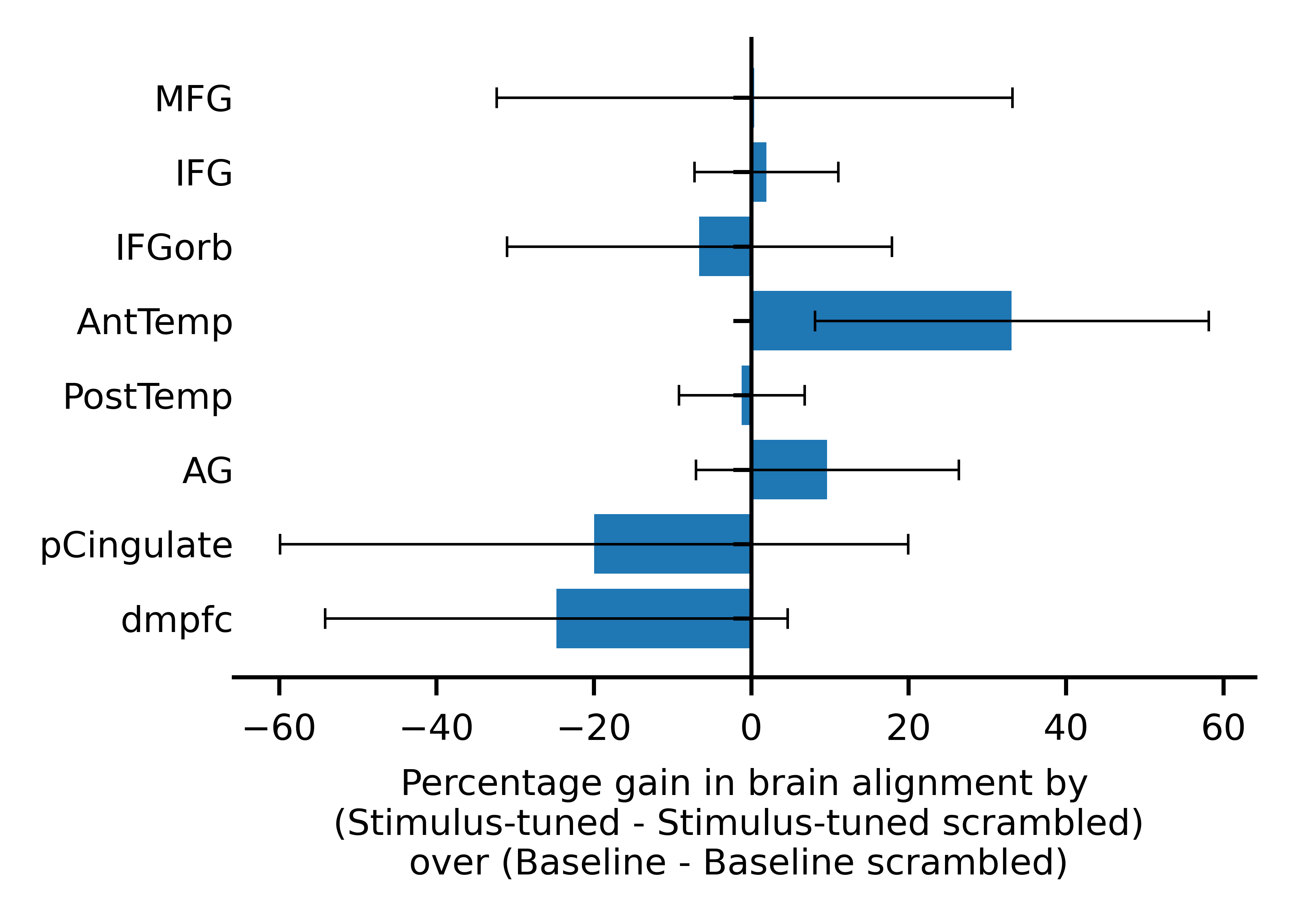}
\caption{Impact of the scrambling perturbation on the stimulus-tuned model versus its impact on the baseline model for GPT-2-small model. We show the median percentage gain as well as the median absolute deviation across participants by (stimulus-tuned - stimulus-tuned scrambled) over (baseline - baseline scrambled) in language regions. We include only voxels with estimated noise ceiling values >0.05.}
  \label{fig:roi_final_gpt2_small}
\end{figure}

\begin{figure}[H]
    \centering
    \includegraphics[width=\textwidth]{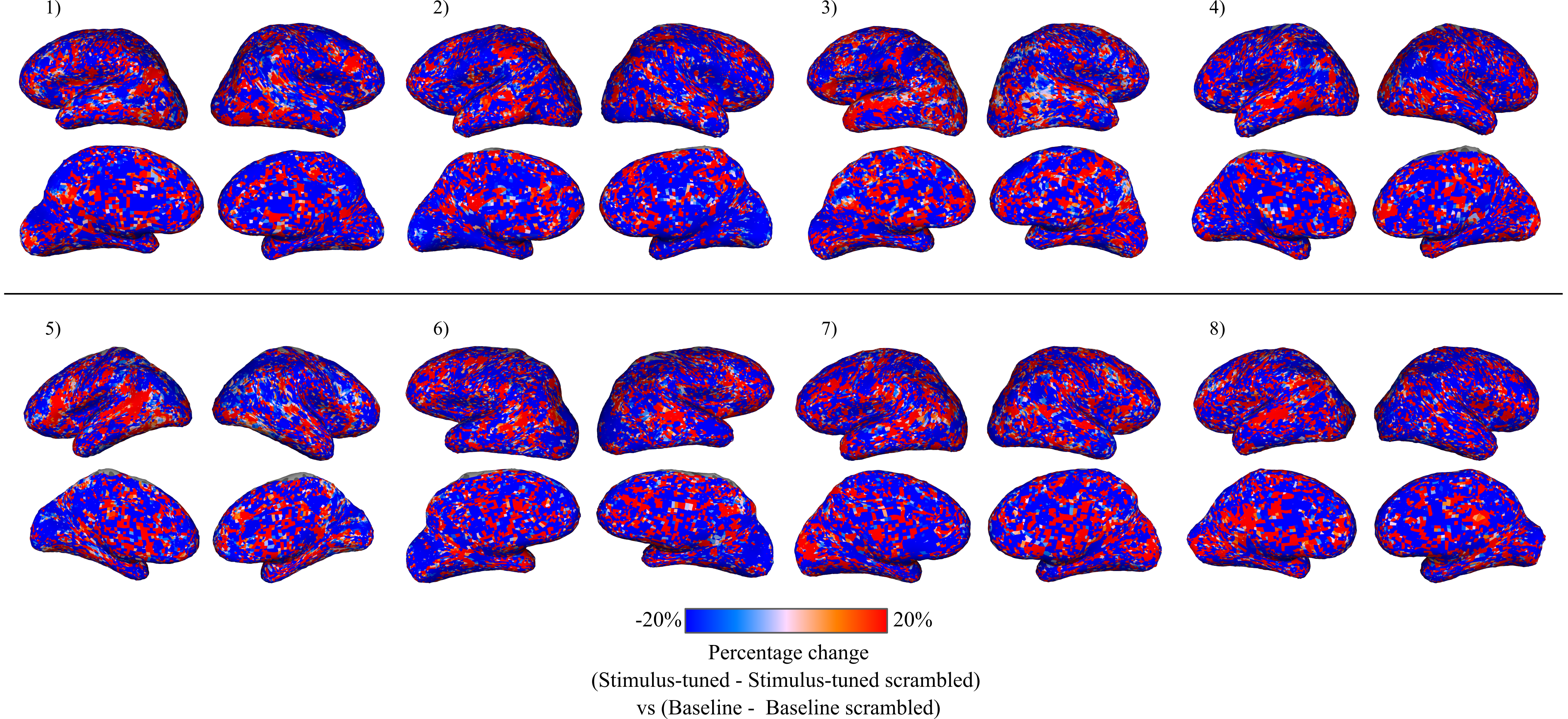}
    \caption{Voxel-wise brain alignment for each participant from contrast that controls for the effect of next-word prediction and word-level information on brain alignment: (stimulus-tuned - stimulus-tuned scrambled) vs (baseline - baseline scrambled). Voxels that appear in red are better predicted by the stimulus-tuned model, even when accounting for next-word prediction and word-level information. Voxels that appear in blue are better predicted by the baseline model. Despite some variation across participants, several language regions appear in red. We quantify these observations in Figure \ref{fig:roi_final_gpt2_small}.}
    \label{fig:appendix_final_contrast_gpt2_small}
\end{figure}

\section{GPT-2-Distilled}\label{appendix:gpt2distilled}

\begin{figure}[H]
\centering
\includegraphics[width=0.5\textwidth]{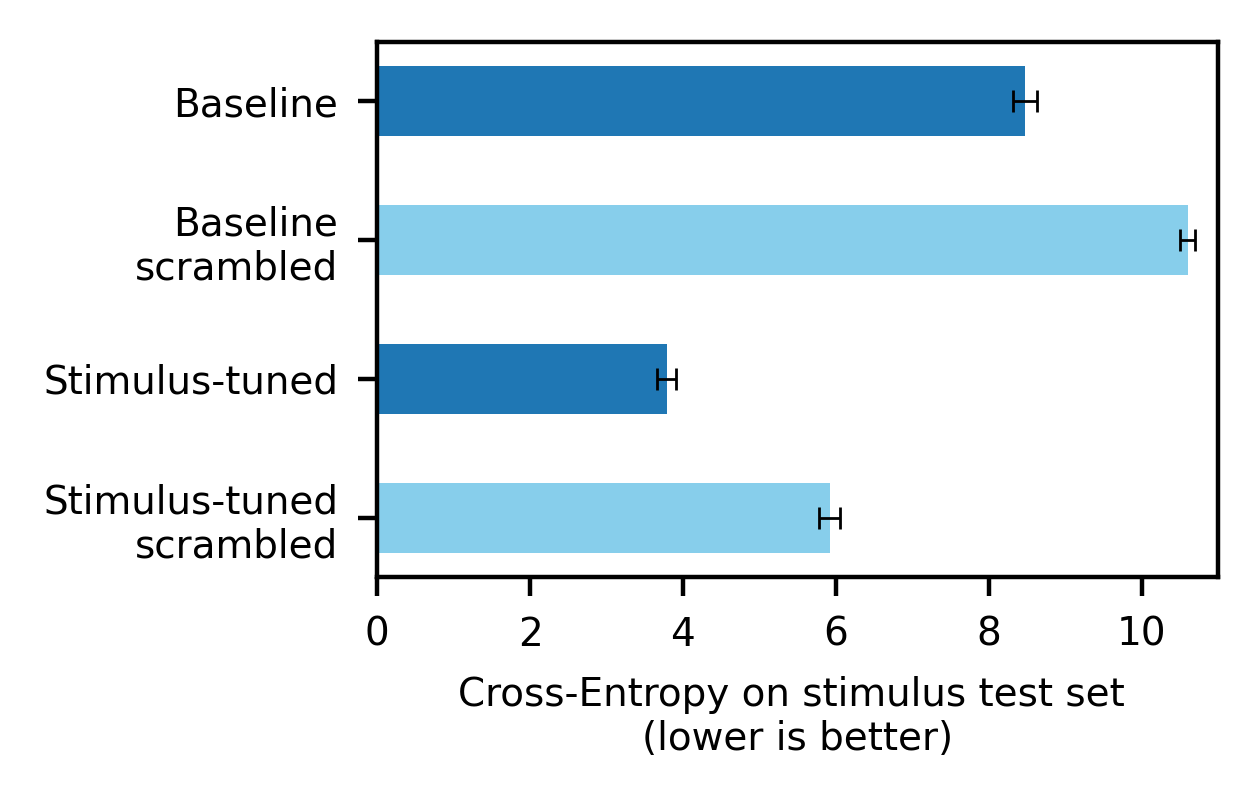}
\caption{Performances of the GPT-2-distilled baseline and perturbed models at next-word prediction averaged across runs with standard deviation.}
  \label{fig:lm_gpt2_distill}
\end{figure}

\begin{figure}[H]
    \centering
    \includegraphics[width=0.95\textwidth]{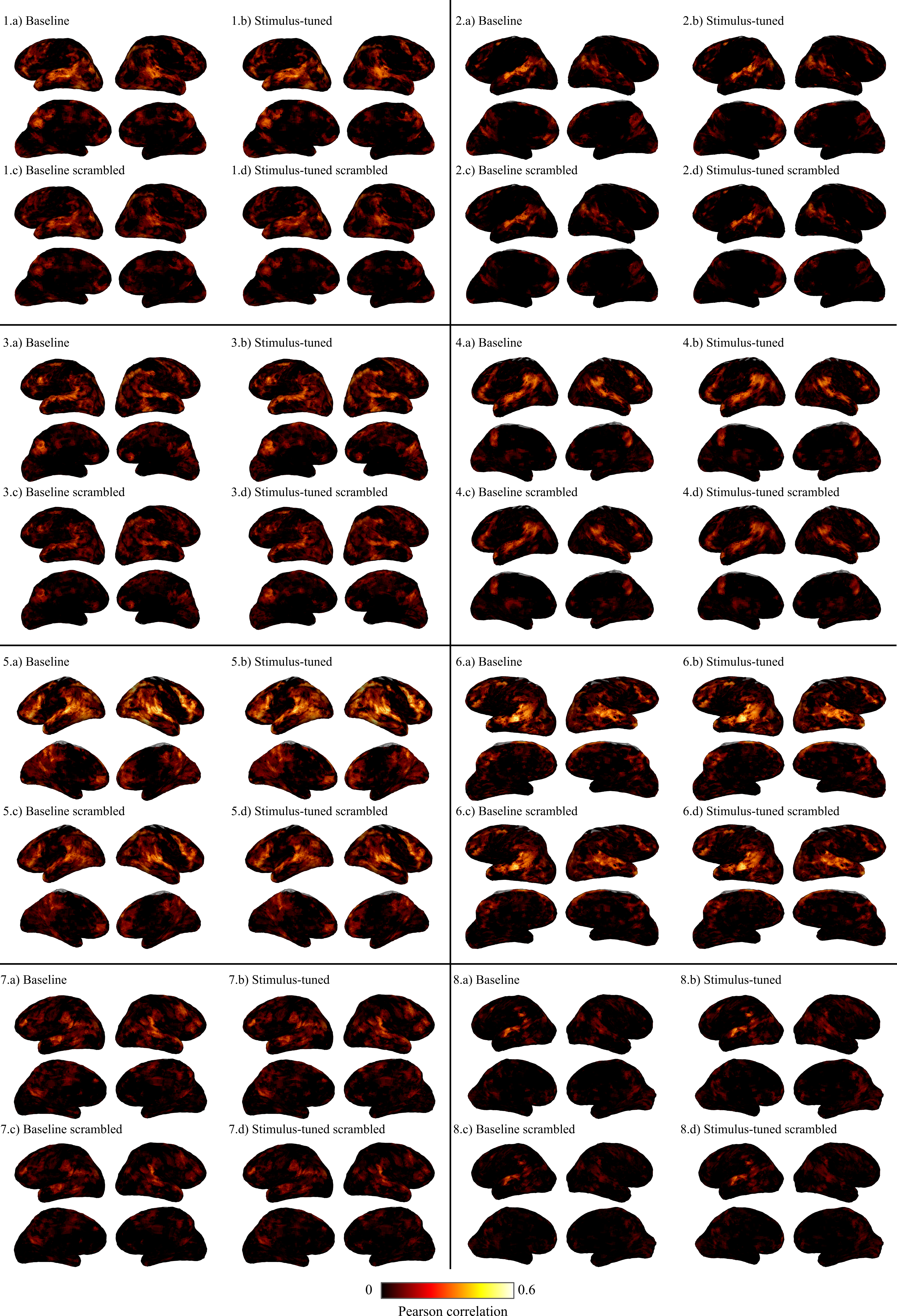}
  \caption{Performances of the GPT-2-distilled baseline and perturbed models of all participants at the brain alignment task. Stimulus-tuning improves the brain alignment (stimulus-tuned in (.b) vs baseline in (.a)) for almost all participants. In contrast, scrambling reduces the brain alignment (baseline in (.a) vs baseline scrambled in (.c)). Despite the reduction in alignment due to the scrambling perturbation, all four models (.a,.b,.c,.d) exhibit alignment in language processing regions.}
  \label{fig:appendix_all_subjects_gpt2_distill}
\end{figure}

\begin{figure}[H]
\centering
\includegraphics[width=0.5\textwidth]{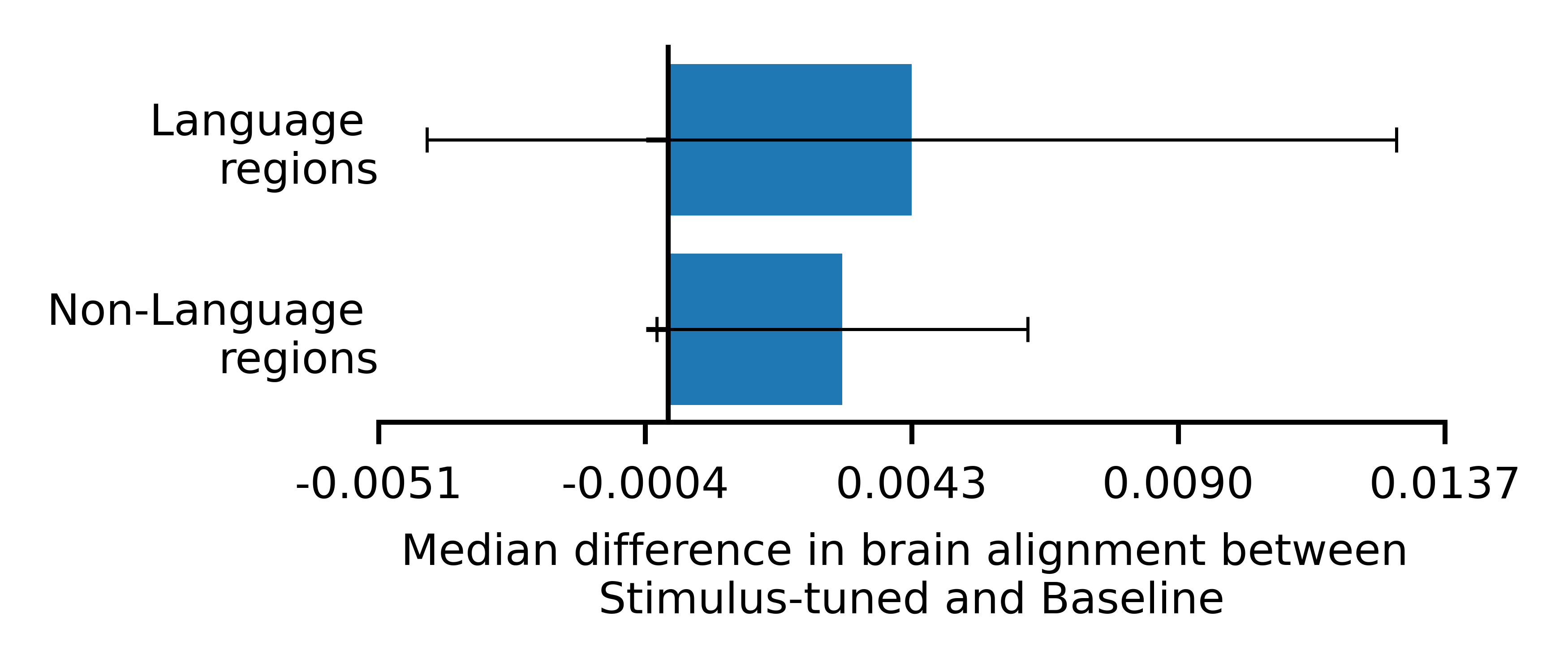}
\caption{Median difference in brain alignment between GPT-2-distilled stimulus-tuned and baseline models. We display the median difference in language and non-language regions and the median absolute deviation across the 8 participants.}
  \label{fig:whole_text_vs_base_gpt2_distill}
\end{figure}

\begin{figure}[H]
\centering
\includegraphics[width=0.5\textwidth]{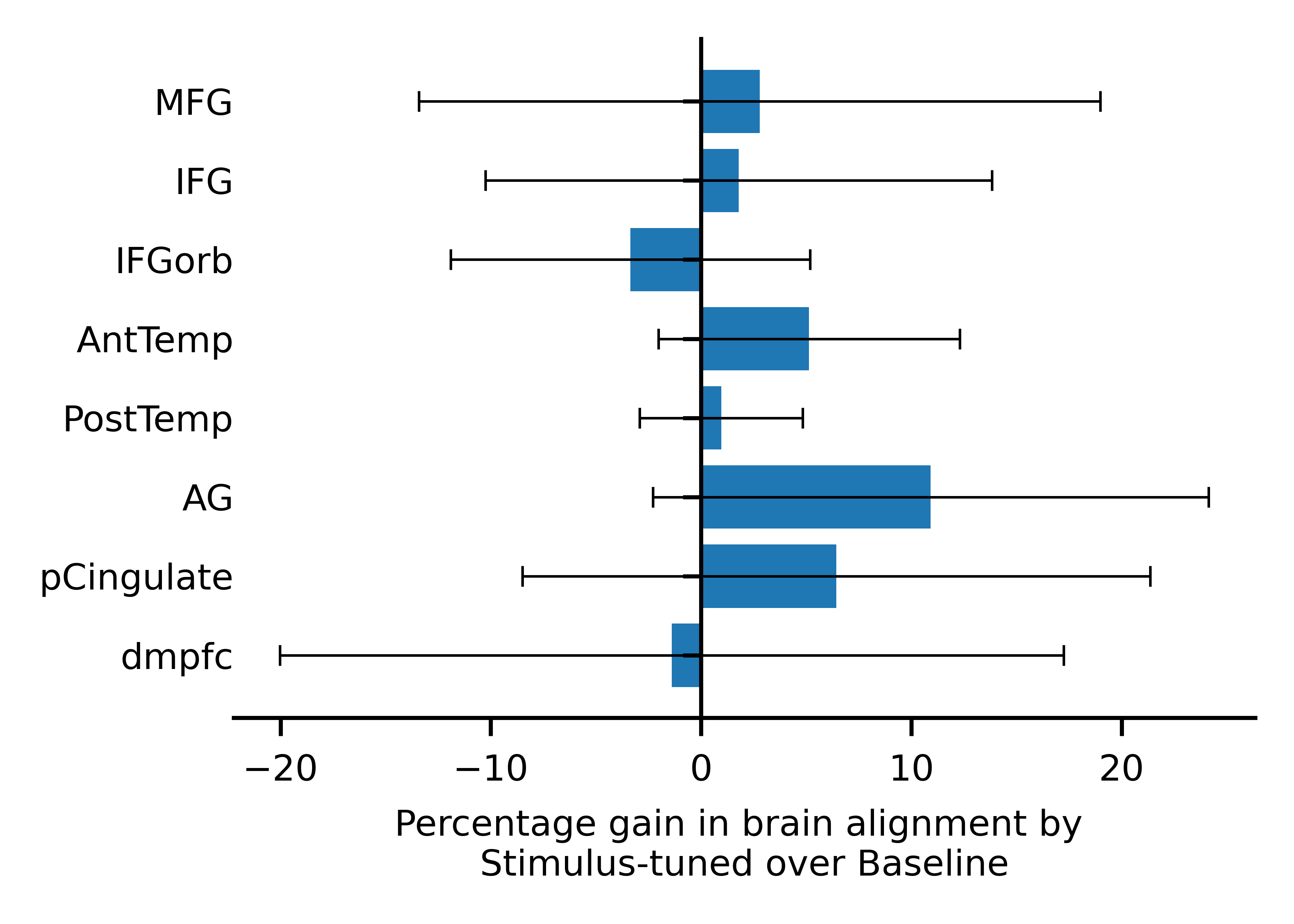}
\caption{Impact of the stimulus-tuning perturbation on the brain alignment of the GPT-2-distilled baseline model. We show the median percentage gain as well as the median absolute deviation across participants. We include only voxels with estimated noise ceiling values >0.05.}
  \label{fig:roi_text_vs_base_gpt2_distill}
\end{figure}

\begin{figure}[H]
\centering
\includegraphics[width=0.5\textwidth]{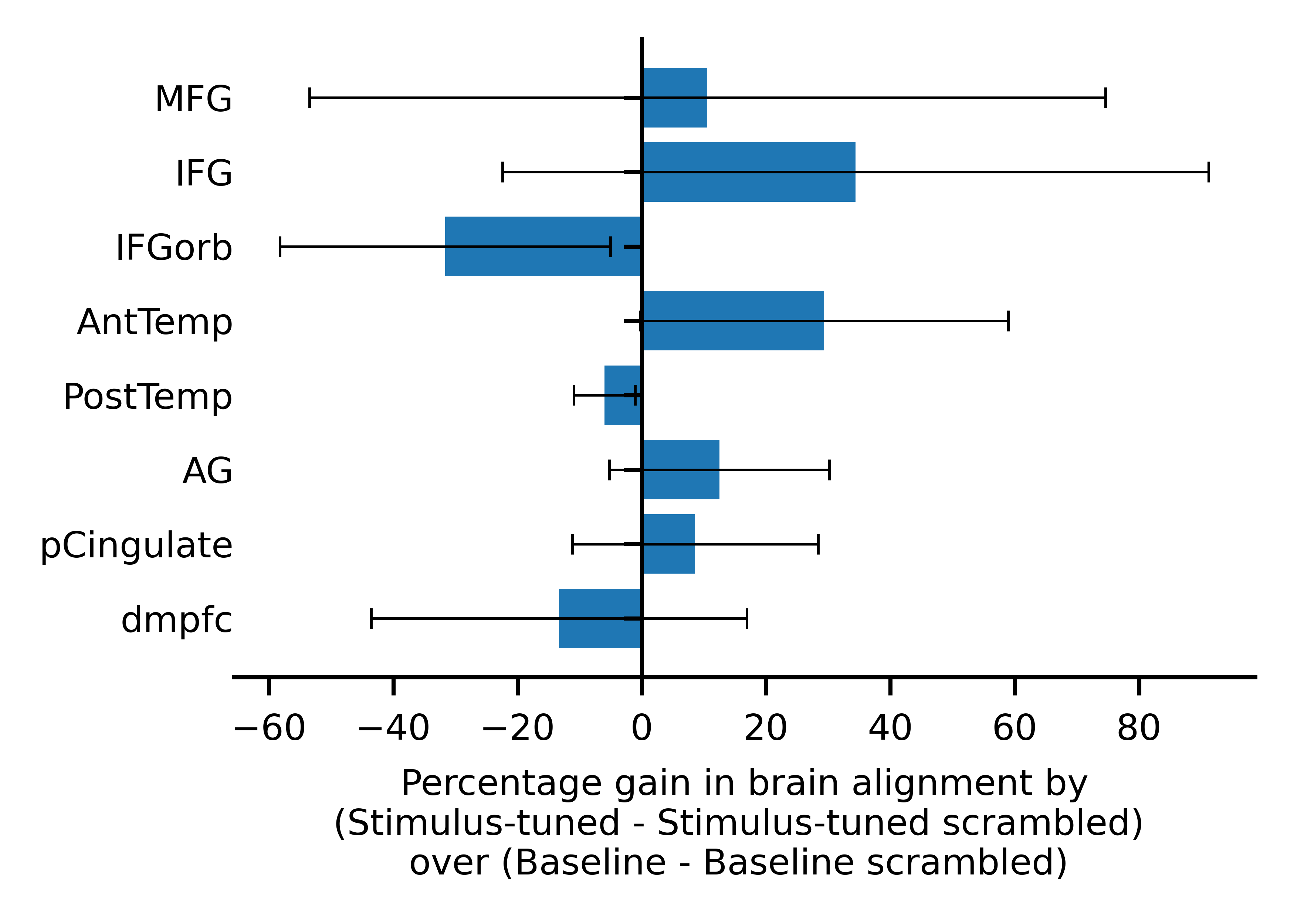}
\caption{Impact of the scrambling perturbation on the stimulus-tuned model versus its impact on the baseline model for GPT-2-distilled model. We show the median percentage gain by (stimulus-tuned - stimulus-tuned scrambled) over (baseline - baseline scrambled) as well as the median absolute deviation across participants in language regions. We include only voxels with estimated noise ceiling values >0.05.}
  \label{fig:roi_final_gpt2_distill}
\end{figure}

\begin{figure}[H]
    \centering
    \includegraphics[width=\textwidth]{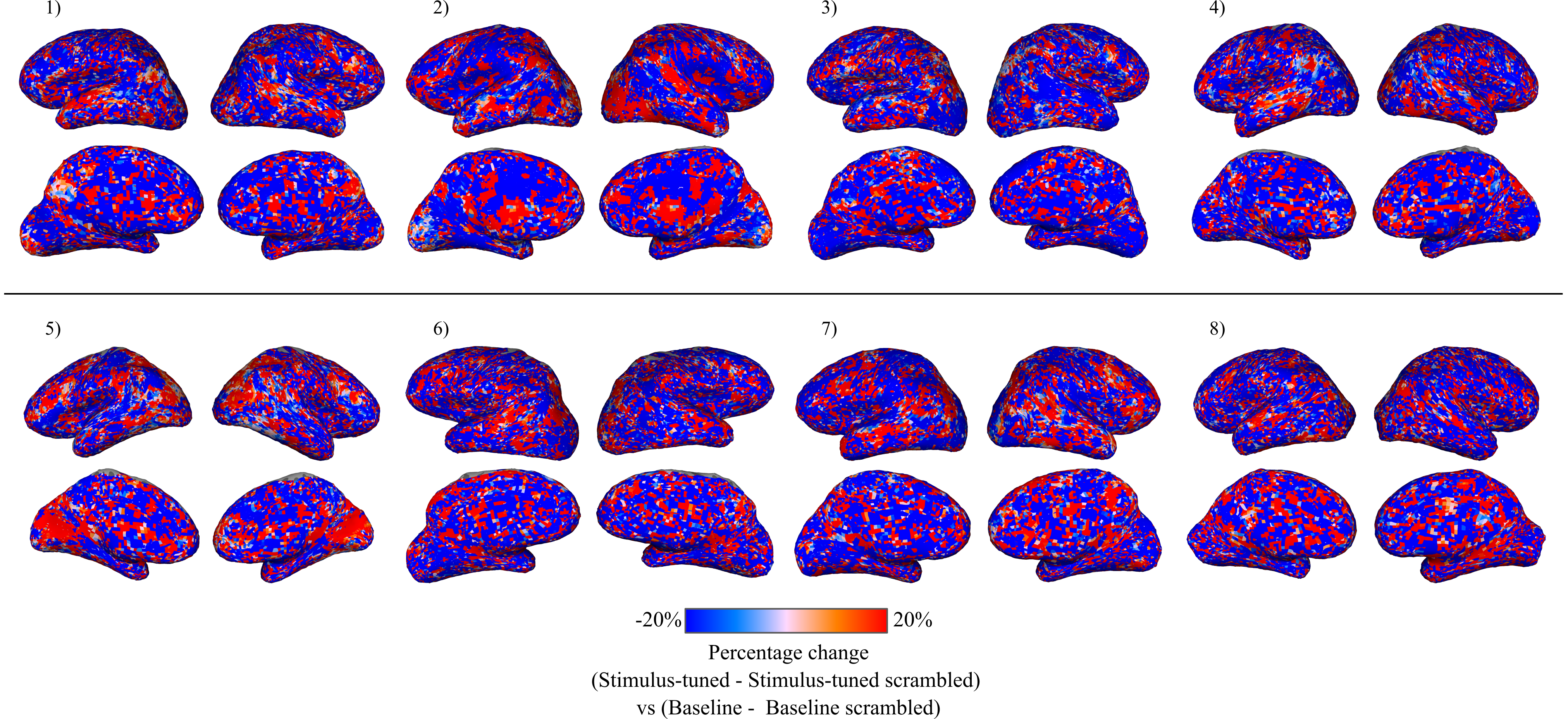}
    \caption{Voxel-wise brain alignment for each participant from contrast that controls for the effect of next-word prediction and word-level information on brain alignment: (stimulus-tuned - stimulus-tuned scrambled) vs (baseline - baseline scrambled) for GPT-2-distill. Voxels that appear in red are better predicted by the stimulus-tuned model, even when accounting for next-word prediction and word-level information. Voxels that appear in blue are better predicted by the baseline model. Despite some variation across participants, several language regions appear in red. We quantify these observations in Figure \ref{fig:roi_final_gpt2_distill}.}
    \label{fig:final_contrast_gpt2_distill}
\end{figure}

\section{GPT-2-Medium}\label{appendix:gpt2medium}

\begin{figure}[H]
\centering
\includegraphics[width=0.5\textwidth]{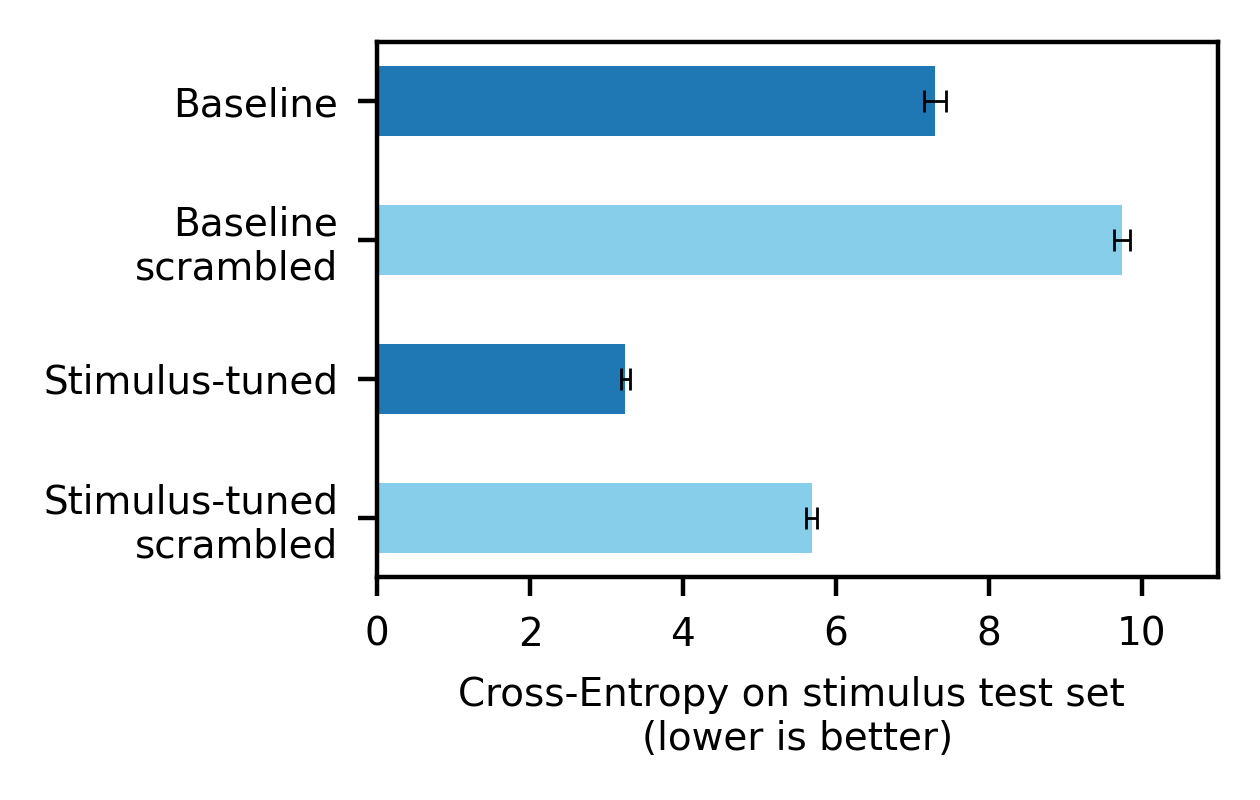}
\caption{Performances of the GPT-2-medium baseline and perturbed models at next-word prediction averaged across runs with standard deviation.}
  \label{fig:lm_gpt2_medium}
\end{figure}

\begin{figure}[H]
    \centering
    \includegraphics[width=0.95\textwidth]{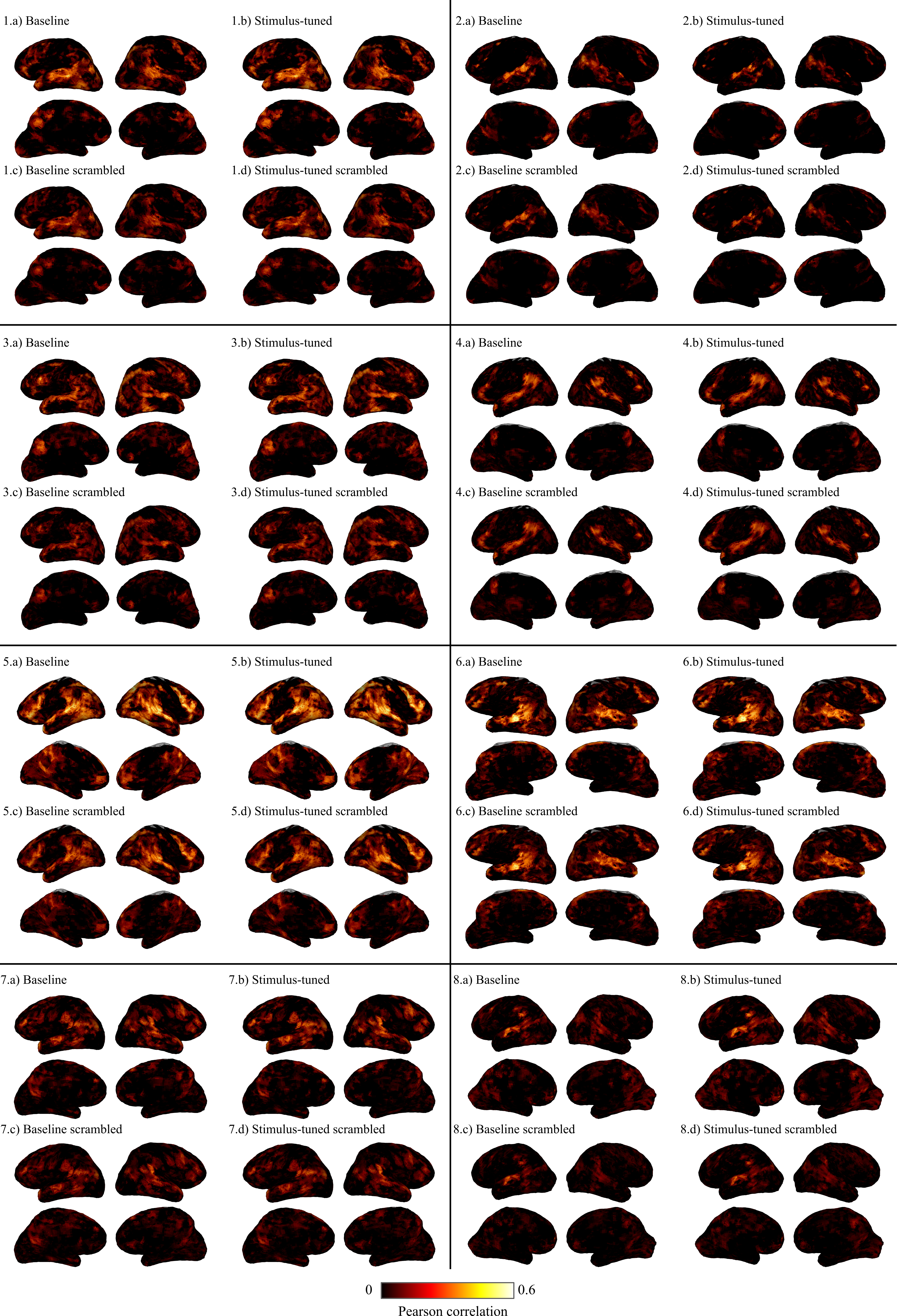}
  \caption{Performances of the GPT-2-medium baseline and perturbed models of all participants at the brain alignment task. Stimulus-tuning improves the brain alignment (stimulus-tuned in (.b) vs baseline in (.a)) for almost all participants. In contrast, scrambling reduces the brain alignment (baseline in (.a) vs baseline scrambled in (.c)). Despite the reduction in alignment due to the scrambling perturbation, all four models (.a,.b,.c,.d) exhibit significant alignment in language processing regions.}
  \label{fig:appendix_all_subjects_gpt2_medium}
\end{figure}

\begin{figure}[H]
\centering
\includegraphics[width=0.5\textwidth]{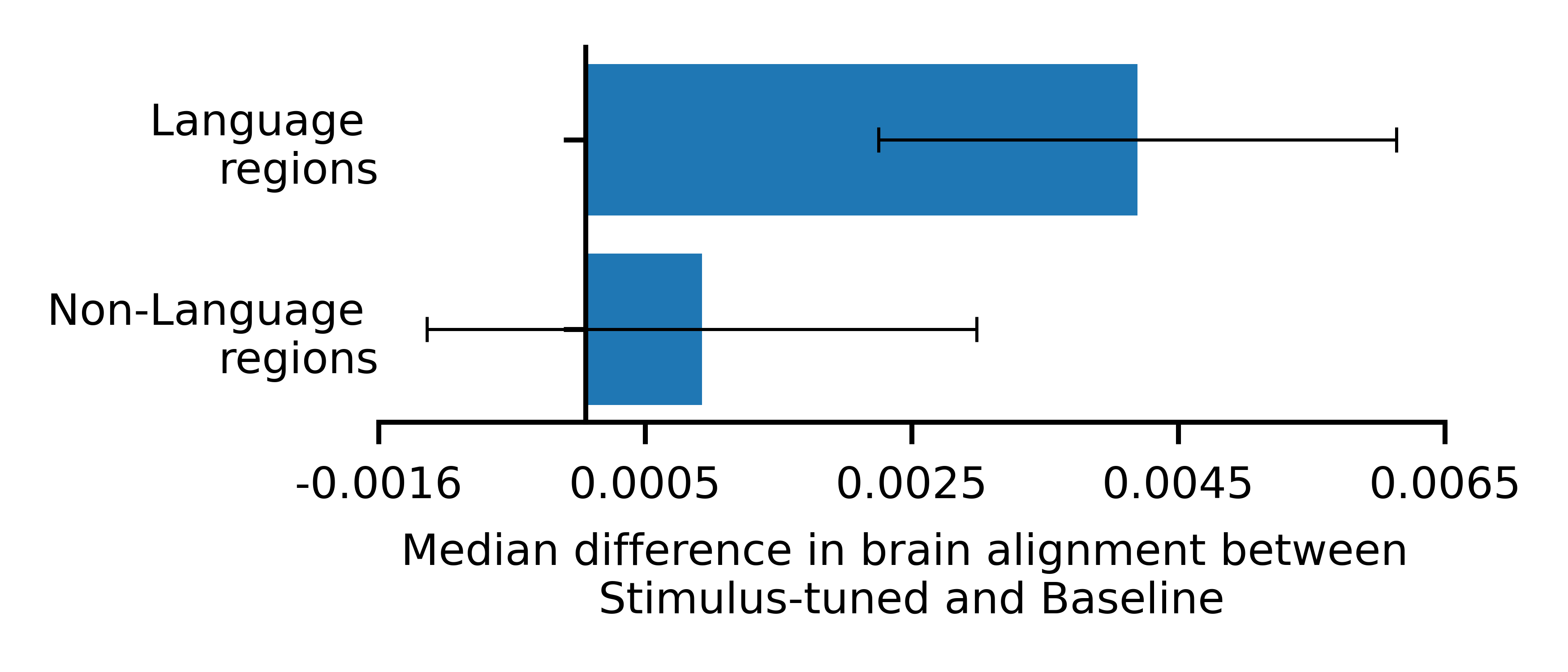}
\caption{Median difference in brain alignment between GPT-2-median stimulus-tuned and baseline models. We display the median difference in language and non-language regions and the median absolute deviation across the 8 participants.}
  \label{fig:whole_text_vs_base_gpt2_medium}
\end{figure}

\begin{figure}[H]
\centering
\includegraphics[width=0.5\textwidth]{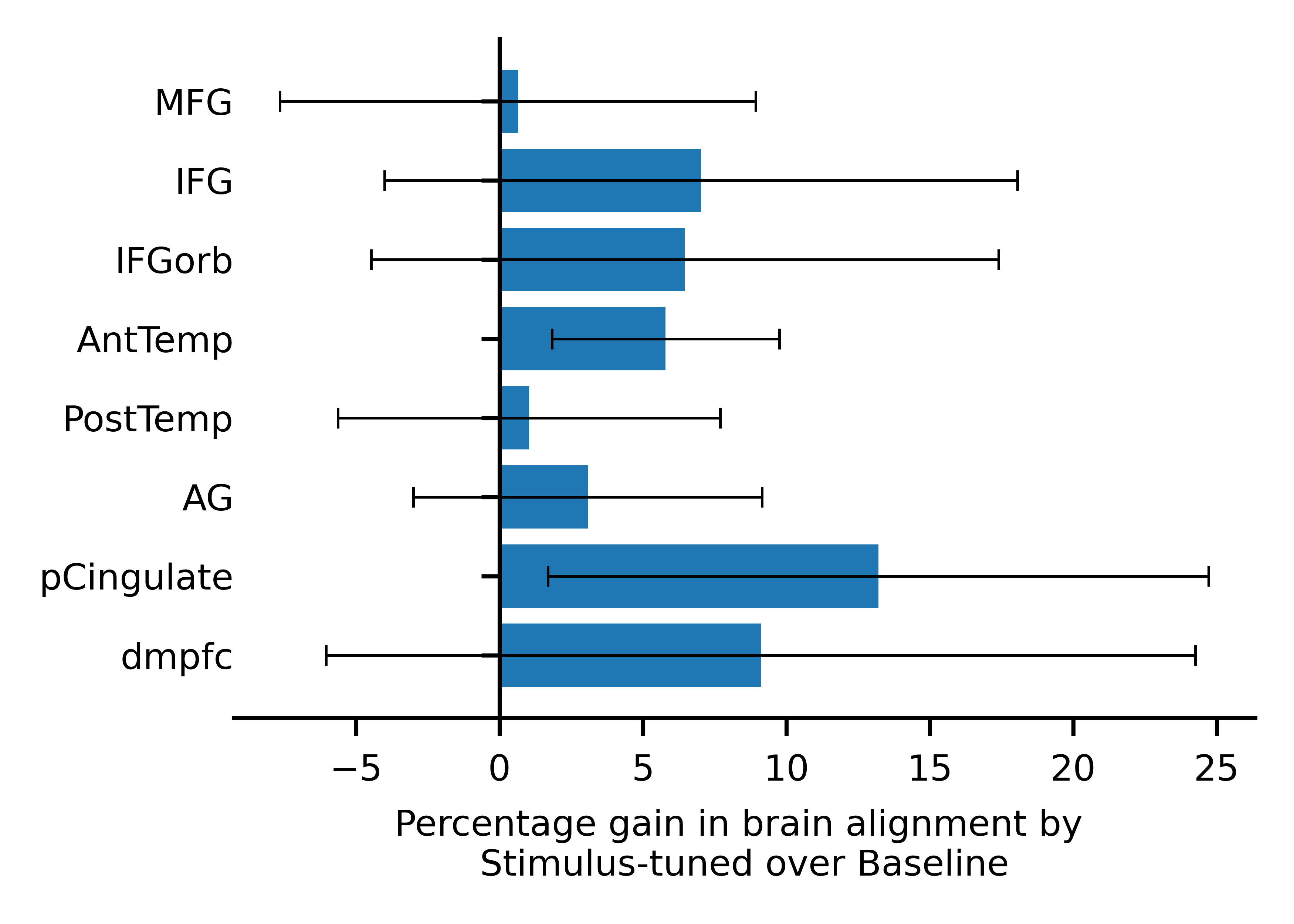}
\caption{Impact of the stimulus-tuning perturbation on the brain alignment of the GPT-2-medium baseline model. We show the median percentage gain as well as the median absolute deviation across participants. We include only voxels with estimated noise ceiling values >0.05.}
  \label{fig:roi_text_vs_base_gpt2_medium}
\end{figure}

\begin{figure}[H]
\centering
\includegraphics[width=0.5\textwidth]{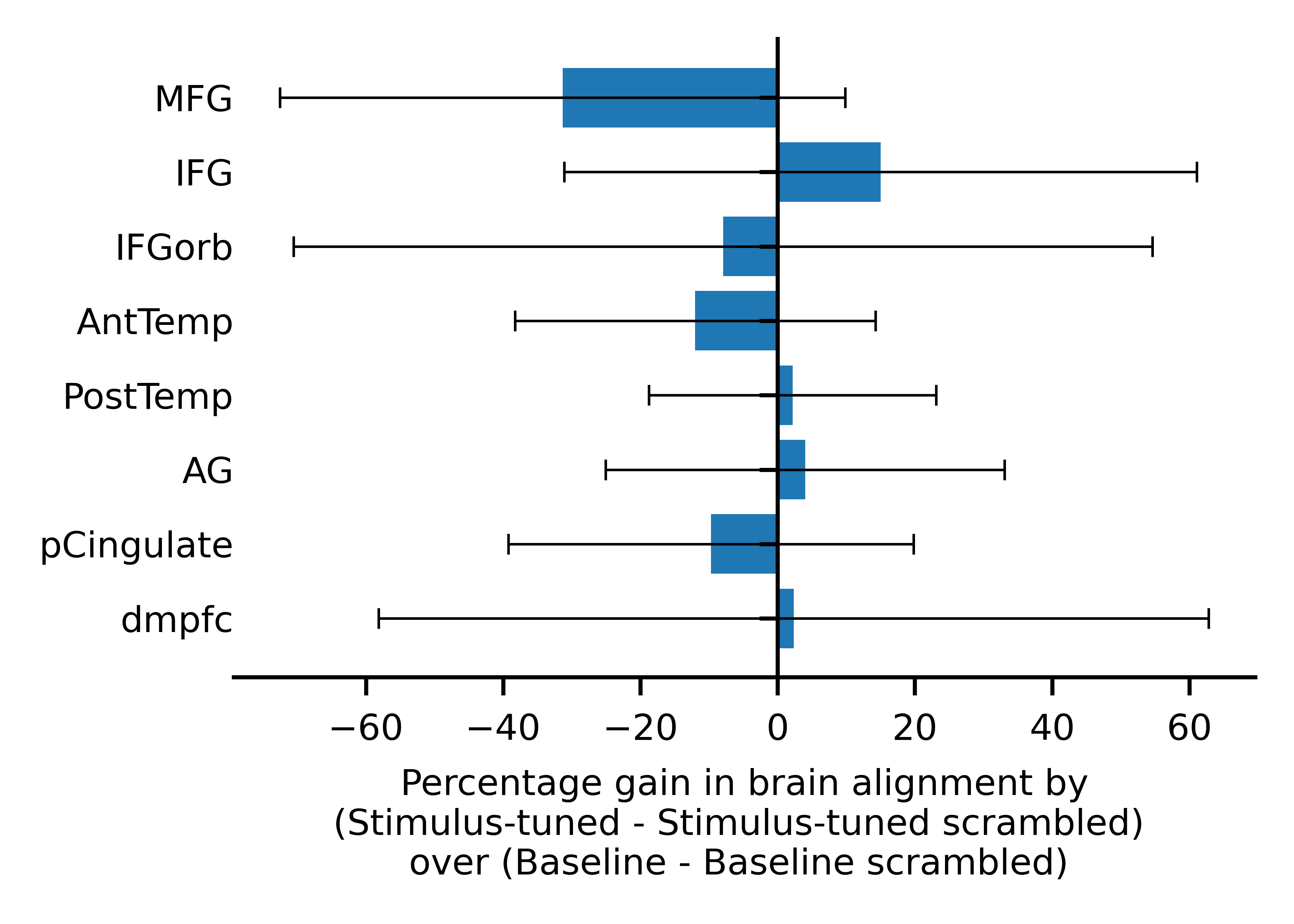}
\caption{Impact of the scrambling perturbation on the stimulus-tuned model versus its impact on the baseline model for GPT-2-medium model. We show the median percentage gain by (stimulus-tuned - stimulus-tuned scrambled) over (baseline - baseline scrambled) as well as the median absolute deviation across participants in language regions. We include only voxels with estimated noise ceiling values >0.05.}
  \label{fig:roi_final_gpt2_medium}
\end{figure}

\begin{figure}[H]
    \centering
    \includegraphics[width=\textwidth]{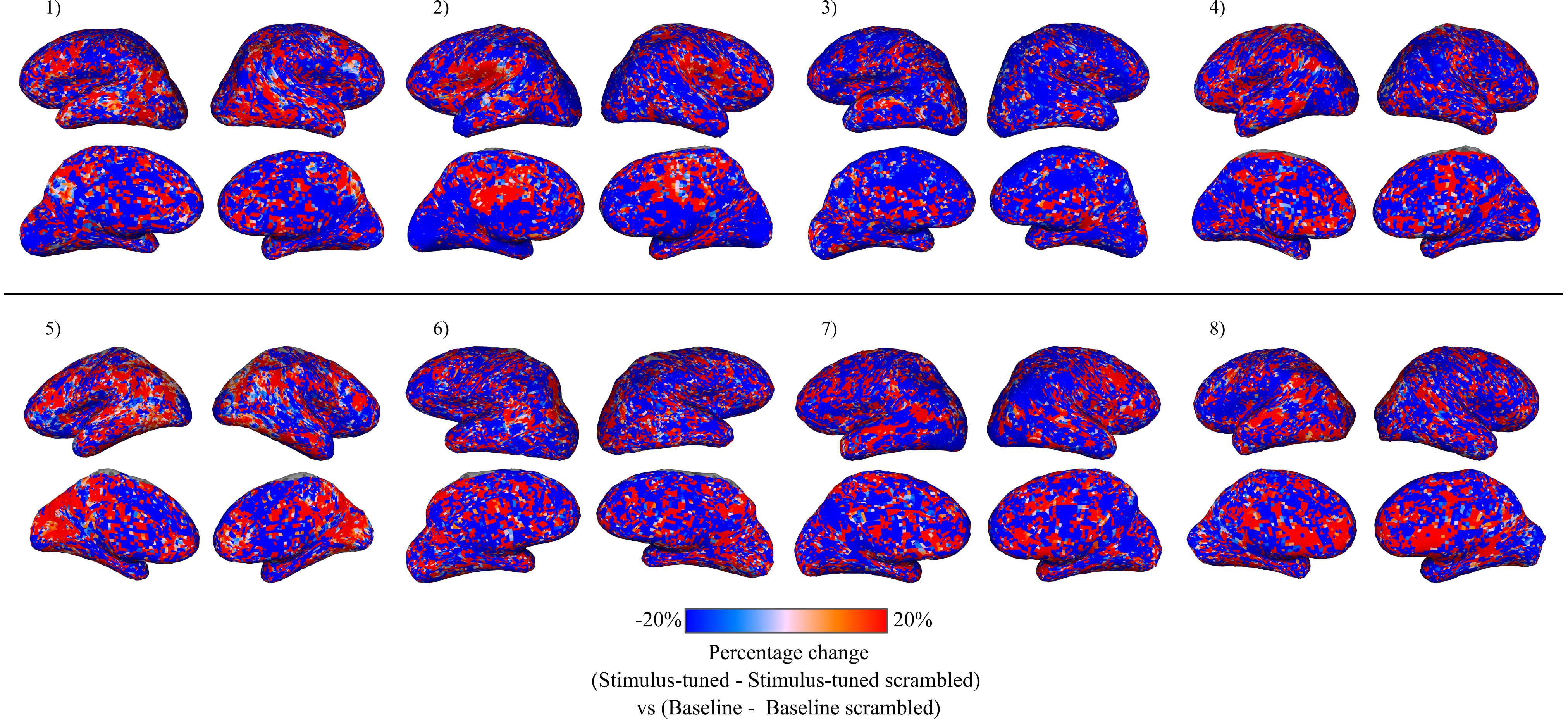}
    \caption{Voxel-wise brain alignment for each participant from contrast that controls for the effect of next-word prediction and word-level information on brain alignment: (stimulus-tuned - stimulus-tuned scrambled) vs (baseline - baseline scrambled) for GPT-2-medium. Voxels that appear in red are better predicted by the stimulus-tuned model, even when accounting for next-word prediction and word-level information. Voxels that appear in blue are better predicted by the baseline model. Despite some variation across participants, several language regions appear in red. We quantify these observations in Figure \ref{fig:roi_final_gpt2_medium}.}
    \label{fig:final_contrast_gpt2_medium}
\end{figure}

\end{document}

%% file: contrasts.tex
\textbf{Baseline $-$ Baseline scrambled:} 
We first consider the contrast of brain alignment between the baseline model and its scrambled counterpart. Any change in brain alignment between the two can be due to changes in word-level information, next-word prediction, or other factors, such as multi-word information:
\begin{equation}
    \Delta_{}^{base}=\Delta_{WL}^{base} + \Delta_{NWP}^{base} + \Delta_{*}^{base},
    \label{eq:base}
\end{equation}
where $\Delta_{}^{base}$ is the change in brain alignment of the baseline model due to scrambling, $\Delta_{WL}^{base}$ is the change in alignment related to differences in word-level information, $\Delta_{NWP}^{base}$ is the change in alignment related to differences in next-word prediction, and $\Delta_{*}^{base}$ is the change in alignment related to other factors, such as multi-word information. By definition, word-level information is not affected by context, so perturbing the order of words in the input does not affect word-level information. Therefore, $\Delta_{WL}^{base}=0$ which simplifies Eq.~\ref{eq:base}:
\begin{equation}
    \Delta_{}^{base}=\Delta_{NWP}^{base} + \Delta_{*}^{base}.
    \label{eq:base-simple}
\end{equation}

\begin{center}
    \textbf{(Stimulus-tuned$-$Stimulus-tuned scrambled)}
\end{center}{}

\begin{center}
    \textbf{$-$ (Baseline$-$Baseline scrambled):}
\end{center}{}

Any residual brain alignment in the previous contrast (i.e. $\Delta_{}^{base}$) may still be due to differences in next-word prediction information ($\Delta_{NWP}^{base}$). Therefore, while the previous contrast controls for the effect of word-level information on brain alignment, it is not able to additionally control for the effect of next-word prediction. To control for the next-word prediction effect, we designed a second-level contrast i.e.  (stimulus-tuned $-$ stimulus-tuned scrambled) vs (baseline $-$ baseline scrambled) ($\Delta_{}^{stim}$ - $\Delta_{}^{base}$). 
Similarly to the baseline model,  we can show that $\Delta_{}^{stim}=\Delta_{NWP}^{stim} + \Delta_{*}^{stim}$. Therefore, the residual brain alignment of the proposed second-level contrast is as follows: 
\begin{align}
\label{eq:second-contrast}
\Delta_{}^{stim}-\Delta_{}^{base}= & \Delta_{NWP}^{stim}+\Delta_{*}^{stim} \\ & -\Delta_{NWP}^{base}-\Delta_{*}^{base}. \notag 
\end{align}
To control for the next-word prediction, we specifically select a fine-tuning checkpoint of the stimulus-tuned model such that the change in next-word prediction performance due to scrambling is similar to that of the baseline (i.e. $\delta^{stim}\approx \delta^{base}$, where $\delta$ signifies the change in next-word prediction between a model and its scrambled counterpart). Therefore, 
$f(\delta^{stim})-f(\delta^{base})\approx 0$, when $f$ is a linear function. Previous work has shown a linear relationship $f$ between next-word prediction and brain alignment \cite{schrimpf2021neural}, which also holds with high correlation in our setting ($0.61$ Pearson correlation, see Appendix \ref{appendix:linear_relation}). Therefore:
\begin{equation}
    \Delta_{NWP}^{stim} - \Delta_{NWP}^{base}\approx 0.
    \label{eq:nwp}
\end{equation}
Combining Eq. \ref{eq:second-contrast} and Eq. \ref{eq:nwp}, we see:
\begin{equation}
    \Delta_{}^{stim}-\Delta_{}^{base}\approx \Delta_{*}^{stim} -\Delta_{*}^{base}.
    \label{eq:final}
\end{equation}
Therefore, if any brain alignment remains after this second-level contrast, it must be due to factors beyond next-word prediction and word-level information, such as multi-word information.